\documentclass[11pt]{article}
\usepackage[]{ACL2023}

\usepackage{times}
\usepackage{latexsym}

\usepackage[T1]{fontenc}

\usepackage[utf8]{inputenc}

\usepackage{microtype}

\usepackage{inconsolata}

\usepackage{amsmath}
\usepackage{amssymb}
\usepackage{graphicx}
\usepackage{todonotes,soul}
\DeclareMathOperator*{\argmax}{argmax} 
\DeclareMathOperator*{\argmin}{argmin} 
\usepackage{subcaption}
\usepackage[skip=0.5ex]{subcaption}
\usepackage[above,below]{placeins}  
\usepackage[mathscr]{euscript}
\usepackage{enumitem}

\usepackage[autostyle, english = american]{csquotes}  
\MakeOuterQuote{"}

\usepackage{cjhebrew}

\setlist{
  listparindent=\parindent,
  parsep=0pt,
}

\title{A Statistical Exploration of Text Partition Into Constituents: The Case of the Priestly Source in the Books of Genesis and Exodus}

\author{Gideon Yoffe \\
  \small \texttt{gideon.yoffe@mail.huji.ac.il}\\\And
  Axel B\"uhler \\
   \small \texttt{axel.buhler@unige.ch}\\
   \And
  Nachum Dershowitz\\
  \small \texttt{nachumd@tauex.tau.ac.il}\\
  \AND
  Israel Finkelstein\\
  \small \texttt{fink2@tauex.tau.ac.il}\\
  \And
  Eli Piasetzky\\
  \small \texttt{eip@tauphy.tau.ac.il}\\
  \And
  Thomas R\"omer\\ \small \texttt{thomas.romer@college-de-france.fr}\\
  \AND
  Barak Sober\\
\small \texttt{barak.sober@mail.huji.ac.il}
\\}

\begin{document}
\maketitle
\begin{abstract}

We present a pipeline for a statistical textual exploration, offering a \textbf{stylometry-based explanation and statistical validation of a hypothesized partition of a text}. Given a parameterization of the text, our pipeline: (\textbf{1}) detects literary features yielding the optimal overlap between the hypothesized and unsupervised partitions, (\textbf{2}) performs a hypothesis-testing analysis to quantify the statistical significance of the optimal overlap, while conserving implicit correlations between units of text that are more likely to be grouped, and (\textbf{3}) extracts and quantifies the importance of features most responsible for the classification, estimates their statistical stability and cluster-wise abundance.

We apply our pipeline to the first two books in the Bible, where one stylistic component stands out in the eyes of biblical scholars, namely, the Priestly component. We identify and explore statistically significant stylistic differences between the Priestly and non-Priestly components.

\end{abstract}

\section{Introduction}

It is agreed among scholars that the extant version(s) of the Hebrew Bible is a result of various editorial actions (additions, redaction, and more).
As such, it may be viewed as a literary patchwork quilt -- whose patches differ, for example, by genre, date, and origin -- and the distinction between biblical texts as well as their relation to other ancient texts from the Near East is at the heart of biblical scholarship, theology, ancient Israel studies, and philology. 
Many debates in this field have been ongoing for decades, if not centuries, with no verdict (e.g., the division of the biblical text into its ``original'' constituents; \cite{gunkel1895creation, von2001old, Wellhausen2009}). While several paradigms gained prevalence throughout the evolution of this discipline \citep{Wellhausen2009, vonRad1973, zakovitch1980study}, the jury is still out on many related hypotheses, such that these paradigms are prone to drastic (and occasionally abrupt) changes over time \citep[e.g.,][]{Nicholson2002}.

When scrutinized, the interrelation between various literary features within the text may shed light on its historical context. From it, one may infer the number of authors, time(s) and place(s) of composition, and even the geopolitical, social, and theological setting(s) \citep[e.g.,][]{Wellhausen2009, vonRad1973, knohl2010divine, Rmer2015}.

Such scrutiny thus serves a double purpose: (\textbf{1}) to disambiguate and identify features in the text that are insightful of the lexical sources of the partition \citep[e.g.,][]{givon1991evolution, deGruyter2018, Peursen2019}; (\textbf{2}) to attempt to trace these features to the context of the text's composition \citep[e.g.,][]{Koppel2011, pat2021syntactic}.

Works employing computational methods in text stylometry -- a statistical analysis of differences in literary, lexical, grammatical or orthographic style between genres or authors \citep{holmes1998evolution} -- have been introduced several decades ago \citep{tweedie1996neural, koppel2002automatically, juola2006prototype, Koppel2009, stamatatos2009survey}, with biblical exegesis spurring very early attempts at computerized authorship-identification tasks \citep{radday1970isaiah, radai1985genesis}. Since then, these methods have proved useful also in investigating ancient \citep[e.g.,][]{kestemont2016authenticating, Verma2017, kabala2020computational}, and biblical texts as well \citep{Koppel2011, Dershowitz2014, roorda2015hebrew, Peursen2019}, albeit to a humbler extent. Finally, statistical-learning-based research, which makes headway in an impressively diverse span of disciplines, is taking its first steps in the context of ancient scripts \citep{Murai2013, FaigenbaumGolovin2016, Dhali2017, Popovic2020, FaigenbaumGolovin2020}. In the biblical context,  \citet{Dershowitz2014, dershowitz2015computerized} addressed reproducing hypothesized partitions of various biblical corpora with a computerized approach as well, using features such as orthographic differences and synonyms. In the first work -- the Cochran-Mantel-Haenszel (CMH) test was applied as a means of hypothesis testing, with the null hypothesis that the synonym features are drawn from the same distribution. While descriptive statistics were successfully applied to various classification and attribution tasks of ancient texts (see above), uncertainty quantification has been insufficiently explored in NLP-related context \citep{dror2018hitchhiker}, and in particular -- in that of text stylometry.

In this work, we introduce a novel exploratory text stylometry pipeline, with which we: (\textbf{1}) find a combination of textual parameters that optimizes the agreement between the hypothesized and unsupervised partitions, (\textbf{2}) test the statistical significance of the overlap, (\textbf3) extract features that are important to the classification, the proportion of their importance, and their relative importance in each cluster. Each stage of this analysis was cross-validated (i.e., applied on many randomly-chosen sub-samples of the corpus rather than applied once on the entire corpus) and tested for statistical stability.

To perform (\textbf{2}) in a meaningful way for textual analysis, we had to overcome the fact that label-permutation tests do not consider correlations between units of text, which affect their likelihood of being clustered together. 
This results in unrealistically optimistic $p$-values, as, in fact, the hypothesized partition \textit{does} implicitly consider such correlations (by, e.g., grouping texts of a similar genre, subject, etc.). 
We overcome this by introducing a cyclic label-shift test which preserves the structure of the hypothesized partition, thus conserving the implicit correlations therein. 
Furthermore, we identify literary features that are \textit{responsible} for the clustering, as opposed to intra-cluster feature selection techniques \citep[e.g.,][]{hruschka2005feature, cai2010unsupervised, zhu2015integrated}, which seek to detect significant features within each cluster. This is also a novel approach to text stylometry.

With this pipeline, we examined the hypothetical distinction between texts of Priestly (P) and non-Priestly (nonP) origin in the books of Genesis and Exodus. The Priestly source is the most agreed-upon constituent underlying the Pentateuch (i.e., Torah). The consensus over which texts are associated with P (mainly through semantic analysis) stems from the stylistic and theological distinction from other texts in the Pentateuch, streamlined across texts associated therewith \citep[e.g.,][]{holzinger1893einleitung, knohl2007sanctuary, romer2014call, faust2019world}. Therefore, the distinction between P and nonP texts is considered a benchmark in biblical exegesis.

\section{Methodology} \label{methodology}

\subsection{Data -- Digital Biblical Corpora} \label{methods_digital_corpus}

We use two digital corpora of the Masoretic variant of the Hebrew Bible in (biblical) Hebrew: (\textbf{1}) a version of the Leningrad codex, made freely available by STEPBible.\footnote{\url{https://github.com/STEPBible/STEPBible-Data}} This dataset comes parsed with full morphological and semantic tags for all words, prefixes, and suffixes. From this dataset, we utilize the grammatical representation of the text through phrase-dependent part-of-speech tags (POS). (\textbf{2}) A digitally parsed version of the Biblia Hebraica Stuttgartensia \citep{roorda2018coding} (hereafter BHSA).\footnote{\url{https://etcbc.github.io/bhsa}} In the BHSA database, we consider the lexematic (i.e., words reduced to lexemes) and grammatical representation of the text through POS. The difference between the two POS-wise representations of the text is that (\textbf{1}) encodes additional morphological information within tags, resulting in several hundreds of unique tags, whereas (\textbf{2}) assigns one out of 14 more ``basic'' grammatical tags\footnote{\url{https://etcbc.github.io/bhsa/features/pdp/}} to each word. We refer to the POS-wise representation of (\textbf{1}) and (\textbf{2}) as ``high-res POS'' and ``low-res POS'', respectively.

\subsection{Manual Annotation of Partition} \label{methods_manualPnonP}

From biblical scholars, we receive verse-wise labeling, assigning each verse in the books of Genesis (1533 verses) and Exodus (1213 verses) to one of two categories: P or nonP, made available online\footnote{\url{https://github.com/YoffeG/PnonP}}. Hereafter, we refer to this labeling as ``scholarly labeling''.

While the dating of P texts in the Pentateuch remains an open, heavily-debated question \citep[e.g.,][]{haran1981behind, hurvitz1988dating, giuntoli2015POSt}, there exists a surprising agreement amongst biblical scholars regarding what verses are affiliated to P \citep[e.g.,][]{knohl2007sanctuary, romer2014call, faust2019world}, amounting to an agreement of 96.5\% and 97.3\% between various biblical scholars for the books of Genesis and Exodus, respectively. We describe the computation of these estimates in Appendix \ref{app_consensus}.

\subsection{Text Parameterization} \label{methods_parameterization}

The underlying assumption in this work is that a significant stylistic difference between two texts of a roughly-similar genre (or, indeed, any number of distinct texts) should manifest in simple observables in NLP, such as the utilization of vocabulary (i.e., distribution of words) and grammatical structure.

We consider three parameters whose different combinations emphasize different properties-, and therefore yield different classifications- of the text:

\begin{trivlist}
    \item[~$\bullet$] \textbf{Lexemes, low-, and high-res POS}: we consider three representations of the text: words in lexematic form and low- and high-res POS (see \S\ref{methods_digital_corpus}). This parameter tests the ability to classify the text based on vocabulary or grammatical structure.
    \item[~$\bullet$] \textbf{$n$-gram size}: we consider sequences of consecutive lexemes/POS of different lengths (i.e., $n$-grams). Different sizes of $n$-grams may be reminiscent of different qualities of the text \citep[e.g.,][]{suen1979, cavnar1994n, ahmed2004language, stamatatos2013robustness}. For example, a distinction based on a larger $n$-gram may indicate a semantic difference between texts or the use of longer grammatical modules in the case of POS (e.g., parallelisms in the books of Psalms and Proverbs \citep{berlin1979grammatical}). In contrast, a distinction made based on shorter $n$-grams is indicative of more embedded differences in the use of language \citep[e.g.,][]{wright2014personality, litvinova2015using}. That said, both these examples indicate that a ``false positive'' distinction can be made where there is a difference in genre \citep[e.g.,][]{feldman2009part, tang2015automatic}. This degeneracy requires careful analysis of the resulting clustering or inclusion of genre-specific texts only for the clustering phase.
    \item[~$\bullet$] \textbf{Verse-wise running-window width}: biblical verses have an average length of 25 words. Hence, a single verse may not contain sufficient context that can be robustly classified. This is especially important since our classification is based on statistical properties of features in the text (see \S\ref{methods_embedding}). Therefore, we define a running window parameter, which concatenates consecutive verses into a single super-verse (i.e., a running window of $k$ would turn the $i${th} verse to the sequence of the $i-k:i+k$ verses) to provide additional context.
\end{trivlist}

\subsection{Text Embedding} \label{methods_embedding}

We consider individual verses, or sequences of verses, as the atomic constituents of the text (see \S\ref{methods_parameterization}). We use tf-idf (term frequency divided by document frequency) to encode each verse, assigning a relevance score to each feature therein \citep{aizawa2003information}. Works such as \citet{fabien2020bertaa, marcinczuk2021text} demonstrate that in the absence of a pre-trained neural language model, tf-idf provides an appropriate and often optimal embedding method in tasks of unsupervised classification of texts. For a single combination of an $n$-gram size and running-window width (using either lexemes, low- or high-res POS), the tf-idf embedding yields a single-feature matrix $X \in \mathbb{R}^{n \times d}$, where $n$ is the number of verses and $d$ is the number of unique $n$-grams of rank $n$. 

It is important to note that this work aims to set a benchmark for future endeavors using strictly traditional machinery throughout our analysis. To ensure collaboration with biblical scholars, our methodology allows for full interpretability of the exploration process (see \S\ref{Results_Features}). This has threefold importance: (\textbf{1}) the field of text stylometry, especially that of ancient Hebrew texts, has hitherto been explored statistically and computationally to a limited extent, such that even when utilizing conservative text-embeddings, such as in this work, considerable insight can be gained concerning both the quality of the analysis and the philological question. (\textbf{2}) Obtaining benchmark results using traditional embeddings is a pre-condition for implementing more sophisticated yet convoluted embeddings, such as pre-trained language models \citep[e.g.][]{shmidman2022introducing} or self-trained/calibrated language models \citep{wald2021calibration}, which we intend to apply in future works. (\textbf{3}) Finally, the interdisciplinary nature of this work and our desire to contribute to the field of biblical exegesis (and traditional philology in general) requires our results to be predominantly interpretable, such that they can be subjected to complementary analysis by scholars from the opposite side of the interdisciplinary divide \citep{piotrowski2012nlp}.

\subsection{Clustering} \label{methods_clustering}

We choose the $k$-means algorithm as our clustering tool of choice \cite[Ch.\ 13.2.1]{hastie2009elements} and hardwire the number of clusters to two, according to the hypothesized P/nonP partition. The justification for our choice of this algorithm is its simple loss-optimization procedure, which is vital to our feature importance analysis and is discussed in detail in \S\ref{methods_interpretability}. 

We use the balanced accuracy (BA) score \citep{sokolova2006beyond} for our overlap statistic, a standard measure designed to address asymmetries between cluster sizes.

Due to the stochastic nature of $k$-means \citep{bottou2004stochastic}, every time it is used in this work -- it is run 50 times -- (with different random initialization) -- and the result yielding the smallest $k$-means loss is chosen.

\subsection{Optimizing for Overlap} \label{methods_optimization}

We perform a 2D grid search over a pre-determined range of $n$-gram sizes and running-window widths to find the parameters combination yielding the optimal overlap for low- or high-res POS lexemes. We test the statistical stability of each combination of these parameters (i.e., to ensure that the overlap reached by each combination is statistically significant) by cross-validating the 2D grid search over some number of randomly-chosen sub-sets of verses, from which we derive the average overlap value for each combination of parameters and the standard deviation thereof. We describe the optimization process in detail in Appendix \ref{app_optimization}.

\subsection{Hypothesis Testing and Validating Results} \label{methods_hypothesisTesting}

\begin{figure*}[t]  
\centering
\includegraphics[scale = 0.6]{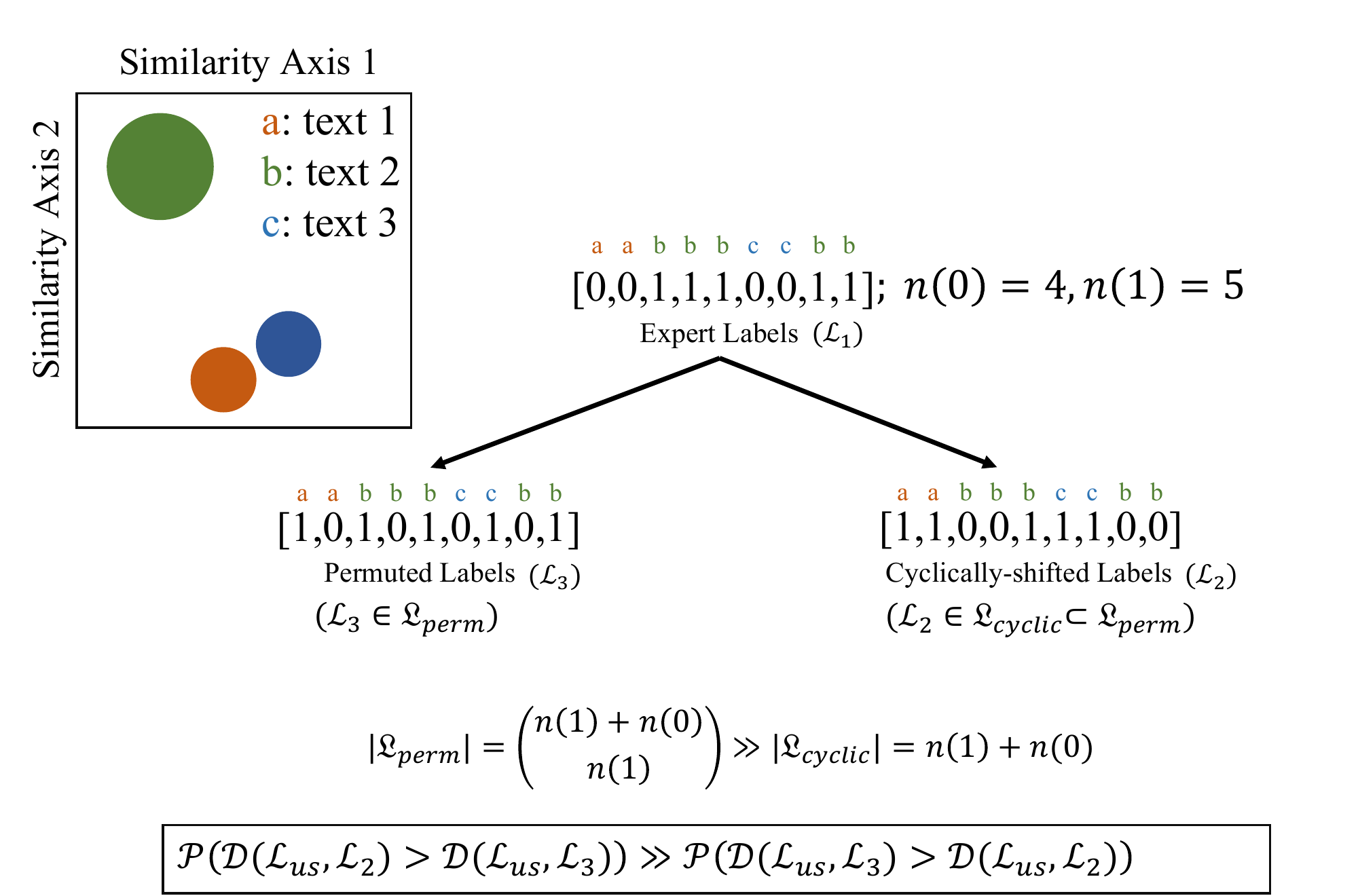}
  \caption{Hypothesis-testing rationale: consider a case of three distinct texts, divided into nine smaller units, whose relative similarity is plotted in the top left diagram. These units of texts are distinguished into two classes by an expert and are labeled as either 0 or 1 ($\mathscr{L}_1$; where $n(0),n(1)$ are the sums of units of text labeled as 0 and 1, respectively). Here, the correlation between adjacent units of text (which is unknown a-priori) is implicitly taken into account, to some degree, by the expert. Consider now the group of all possible permutations ($\mathfrak{L}_{perm}$) of the expert and the group containing all the possible cyclic shifts thereof ($\mathfrak{L}_{cyclic}$). The latter is a sub-group of -- but is much smaller than -- the first ($\mathfrak{L}_{cyclic} \subset \mathfrak{L}_{perm}$). For sufficiently long texts, we argue that the probability ($\mathscr{P}$) of randomly permuting the expert labeling such that it would capture some intrinsic correlations between the units of text ($\mathscr{L}_3$) and would yield a substantial overlap (expressed in the figure as the function $\mathscr{D}$, which in our case is the BA score) with an unsupervised clustering labeling ($\mathscr{L}_{us}$) of the text -- or at least equivalent to that of the expert labeling -- is considerably lower than that of a cyclically-shifted expert labeling ($\mathscr{L}_2$), which, up to a shift, assumes the implicit correlations between units of text manifested in $\mathscr{L}_1$. These, in turn, are more likely to yield equivalent overlap values to $\mathscr{L}_1$. A cyclic-shift test is, therefore, a more prohibitive hypothesis-testing routine than the naive permutation test.}
     \label{Fig_hypothesis_testing_scheme}
\end{figure*}

Through hypothesis testing, we establish the statistical significance of the achieved optimal overlap value between the unsupervised and hypothesized partitions. 
To derive a $p$-value from some empirical null distribution, we consider the assumption that the hypothesized partition, manifested in the scholarly labeling, exhibits a specific formulaic partition of chunks of the text. A formulaic partition, in turn, suggests that verses within each of the P (nonP) blocks are correlated -- a fact that the standard label-permutation test is intrinsically agnostic of, as it permutes labels without considering potential correlations between verses (Fig. \ref{Fig_hypothesis_testing_scheme} left). 
Thus, the null distribution synthesized through a series of permutations would represent an overly-optimistic scenario that does not correspond to any conceivable scenario in text stylometry. 

To remedy this, we devise a more prohibitive statistical test. Instead of permuting the labels to have an arbitrary order, we perform a cyclic shift test of the scholarly labeling (Fig. \ref{Fig_hypothesis_testing_scheme} right). 
This procedure retains the scholarly labels' hypothesized \textit{structure} but shifts them across different verses. We perform as many cyclic shifts as there are labels (i.e., number of verses) in each book by skips of twice the largest running-window width considered in the optimization procedure. For each shift, we perform the parameter optimization procedure (see \S\ref{methods_optimization}), where we now have the shifted scholarly labels instead of the original (``un-shifted'') ones. Thus, we generate a distribution of our statistic under the null hypothesis, from which we derive a $p$-value. 

In Fig.~\ref{Fig_hypothesis_testing_scheme}, we present an intuitive scheme 
where we demonstrate our rationale concerning the hypothesis-testing procedure in text stylometry.

\subsection{Feature Importance and Interpretability of Classification} \label{methods_interpretability}

Given a $k$-means labeling produced for the text, which was embedded according to some combination of parameters (\S\ref{methods_parameterization}), we wish to quantify the importance of individual $n$-grams to the classification, the proportion of their importance, and associate to which cluster they are characteristic of. 

Consider the loss function of the $k$-means algorithm:
\begin{equation*} \label{eq_kmeans}
    \argmin_S\sum _{i = 1}^k |S_i|\cdot var(S_i),
\end{equation*}
where $k=2$ is the number of desired clusters, $S$ is the group of all potential sets of verses, split into $k$ clusters, $|S_i|$ is the size of the $i${th} cluster (i.e., number of verses therein) and $var(S_i)$ is the variance of the $i${th} cluster. That is, the $k$-means aims to minimize intra-class variance.
Equivalently, we could optimize for the \textit{maximization} of the \textit{inter-cluster} variance (i.e., the variance between clusters), given by      
\begin{equation} \label{eq_kmeans_loss}
        \argmax_S\sum _{i \in S_1}\sum_{j\in S_2}  \|{x}_{i} - x_j\|^2
.\end{equation}

Let $D \in \mathbb{R}^{|S_1|\cdot |S_2| \times d}$ denote the matrix of inter-cluster differences whose columns are $D_{\ell}$, which for $i\in \{1,\ldots, |S_1|-1\}$, $ j\in\{1,\ldots |S_2|\}$ and $\ell = i \cdot |S_2| + j$ are defined by
\begin{equation*}
    (D)_\ell = x_i - x_j
.\end{equation*}
Then, applying PCA to $D$ would yield the axis across which Eq. \eqref{eq_kmeans} is optimized as the first \textit{principal component}.  This component represents the axis of maximized variance, and each feature's contribution is given by its corresponding loading. Finally, it can be shown that this principal axis could also be computed by subtracting the centroids of the two clusters.
Therefore, to leverage this observation and extract the features' importance, we perform the following procedure:
\begin{trivlist}
    \item[~$\bullet$] Compute the \textit{principal separating axis} of the two clusters by subtracting the centroid of $S_1$ from that of $S_2$.
    \item[~$\bullet$] The contribution of each $n$-gram feature to the cluster separation is given by its respective loading in the principal separating axis.
    \item[~$\bullet$] Since tf-idf assigns strictly non-negative scores to each feature, the signs of the principal separating axis' loadings allow us to associate the importance of each $n$-gram to a specific cluster (see Appendix \ref{app_features}).
    \item[~$\bullet$] To determine the stability of the importance of $n$-grams across multiple sub-samples of each book, we perform a cross-validation routine (similar to the one performed in \S\ref{methods_optimization}). Explicitly, we perform all the steps listed above for some number of simulations (i.e., randomly sample a sub-set of verses and extract the importance vector) and compute the mean and standard deviation of all simulations.
    \item[~$\bullet$] Finally, we compute the relative uniqueness of each important feature w.r.t.\@ both clusters -- assigning it a score between $-1:1$. A score of 1 ($-1$) indicates that the feature is solely abundant in its associated (opposite) cluster. Thus, a feature's abundance nearer zero indicates that its contribution to the clustering is rather in its \textit{combination} with other features than a standalone indicator. 
\end{trivlist}

\section{Results} \label{results}

We apply a cross-validated optimization analysis to the three representations of the book (\S\ref{methods_optimization}) by performing 2D grid searches for 20 randomly chosen sub-sets of 250 verses for each representation. We perform a cross-validated cyclic hypothesis-testing analysis for the optimal overlap value (\S\ref{methods_hypothesisTesting}) using five randomly-chosen sub-sets of verses, similarly to the above -- and derive a $p$-value. Finally, we perform a cross-validated feature importance analysis for every representation (\S\ref{methods_interpretability}), over 100 randomly-chosen sub-sets of verses, similar to the above. 

\begin{figure*}
\centering
\includegraphics[scale = 0.8]{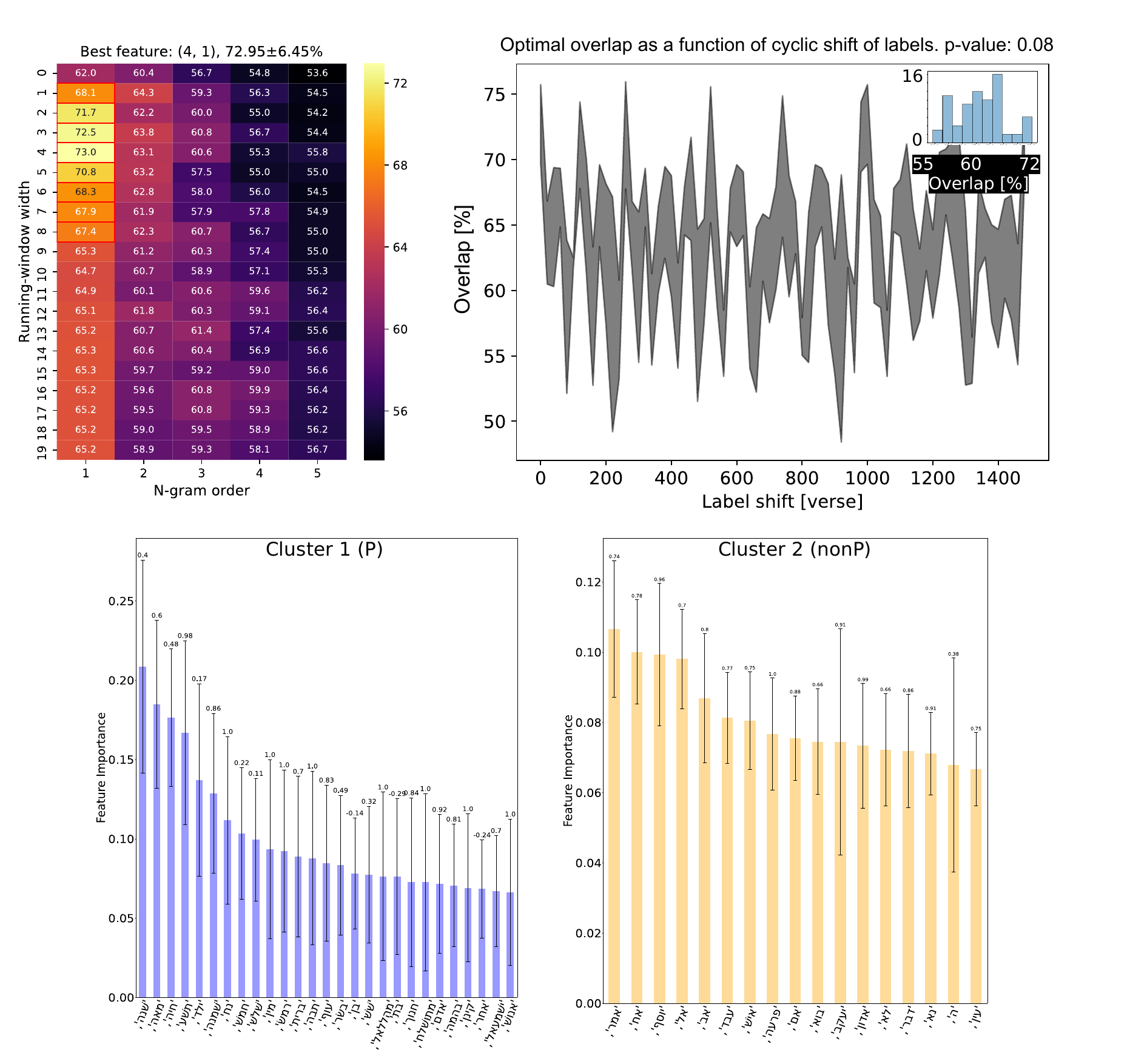}
  \caption{A statistical exploration of the hypothesized partition of the books of Genesis and Exodus into Priestly and non-Priestly constituents: results for the book of Genesis (lexemes representation). \textbf{Upper Left Panel -- Optimization}: Cross-validated grid-search over verse running-window widths and $n$-gram sizes to identify the combination yielding an optimal overlap. The combination, value, and uncertainty of the optimal overlap are plotted on top of the panel, and all combinations whose overlap value is within 1$\sigma$ of the optimal overlap value are marked in red cells. \textbf{Upper Right Panel -- Hypothesis Testing}: Cross-validated hypothesis testing, where we simulate the null hypothesis distribution through a cyclic shift of the hypothesized P/nonP labeling. The $p$-value is measured with respect to the optimal overlap value minus its standard deviation. \textbf{Bottom Panels -- feature importance Analysis}: Cross-validated important feature analysis (running-window 4 and $n$-gram size 1) and their statistical stability, displaying features bearing 75\% of the explained variance. Features in the left and right panels are important to the P-associated and nonP-associated clusters, respectively. The small numbers above each error bar indicate the cluster-wise abundance of the feature.}
     \label{Fig_PnonP_Genesis}
\end{figure*}

In Table \ref{Tab_results}, we list the cross-validated results of each representation for both books and the derived $p$-value. In Fig.\@ \ref{Fig_PnonP_Genesis}, we visualize results for all stages of our analysis applied to the lexematic representation of the book of Genesis. In Appendix \ref{app_results}, we plot the results for all stages of our analyses for both books and discuss them in detail. Appendix \ref{app_disc} presents a detailed biblical-exegetical analysis of our results and an expert's evaluation of our approach. 

\begin{table*}
\small\tabcolsep1pt
    \centering
    \begin{tabular}{c c c c c} 
    \hline\hline
    Book & Opt. overlap (lexemes) & Opt. overlap (low-res POS) & Opt. overlap (high-res POS) & $p$-value \\ 
    \hline
    Genesis & 72.95$\pm$6.45\% (\textit{rw}: 4, $n$: 1) & 65.03$\pm$5.64\% (\textit{rw}: 14, $n$: 1) & 73.96$\pm$2.91\% (\textit{rw}: 4, $n$: 1) & 0.08 (low-res POS) \\
    Exodus & 89.23$\pm$2.53\% (\textit{rw}: 8, $n$: 2) & 88.63$\pm$1.96\% (\textit{rw}: 9, $n$: 4) & 86.53$\pm$2.91\% (\textit{rw}: 6, $n$: 2) & 0.06 (high-res POS) \\

    \hline
    \end{tabular}
    \caption{Cross-validated optimization and hypothesis testing results: for each representation, we list the optimal overlap value, its respective uncertainty, and combination of parameters (\textit{rw} for running-window width and $n$ for $n$-gram size).}
    \label{Tab_results}
\end{table*}

\subsection{Understanding the Discrepancy in Results Between the Books of Genesis and Exodus} \label{results_discrepancyExodusGenesis}

In Table \ref{Tab_results}, we list our optimal overlap values for all textual representations of both books. Notice that there exists a statistically-significant discrepancy between the optimal overlap values between both books that is as follows: 

For Exodus, all three representations reach the same optimal overlap of roughly 88\%. In contrast, in Genesis -- there is a difference between the achieved optimal overlap values for the lexematic and low-res POS representations on the one hand (73\%) and high-res POS representation on the other (65\%). 

\begin{figure}[h]
\centering
\includegraphics[scale = 0.45]{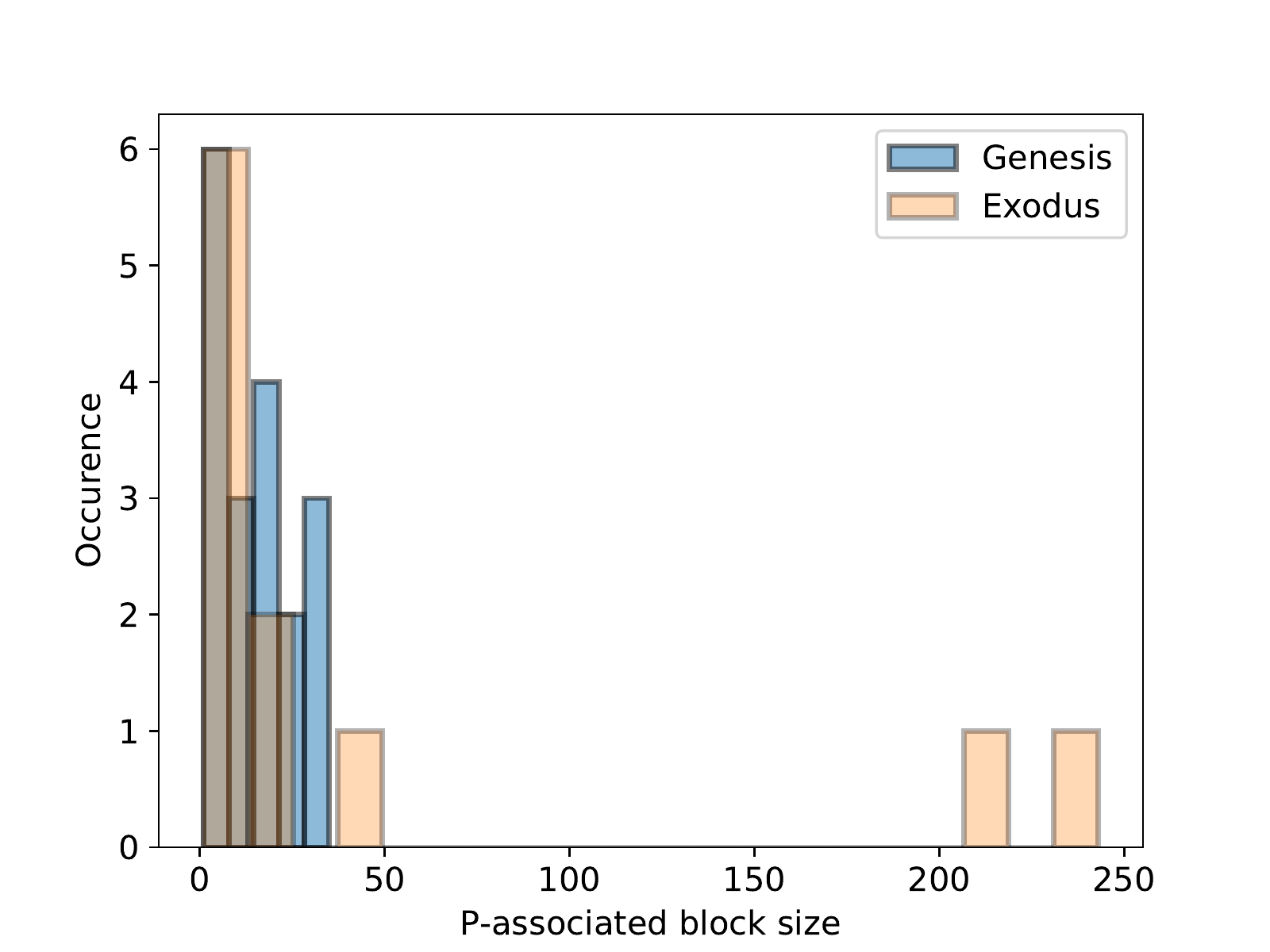}
  \caption{Size distribution histogram of P-associated blocks in the books of Genesis and Exodus.}
     \label{Fig_P-associated_blocks_hist}
\end{figure}

When examining the verses belonging to each cluster when classified with the optimal parameter combination, it is evident that most of the overlapping P-associated verses in Exodus are grouped in two blocks of P-associated of text, spanning 243 and 214 verses. These make considerable outliers in the size distribution of P-associated blocks in both books (Fig.\@ \ref{Fig_P-associated_blocks_hist}), which may be related to the observed discrepancy.

This discrepancy begs two questions: 

\begin{enumerate}

\item Are the linguistic differences between P/nonP -- which may be captured by our analysis -- that are attenuated for shorter sequences of P texts?
\item Does the high overlap in the book of Exodus arise due to an implicit sensitivity to the generic/semantic uniqueness of the two big P-associated blocks rather than a global stylistic difference between P/nonP?
\end{enumerate}

\begin{figure}[h]
\centering
\includegraphics[scale = 0.45]{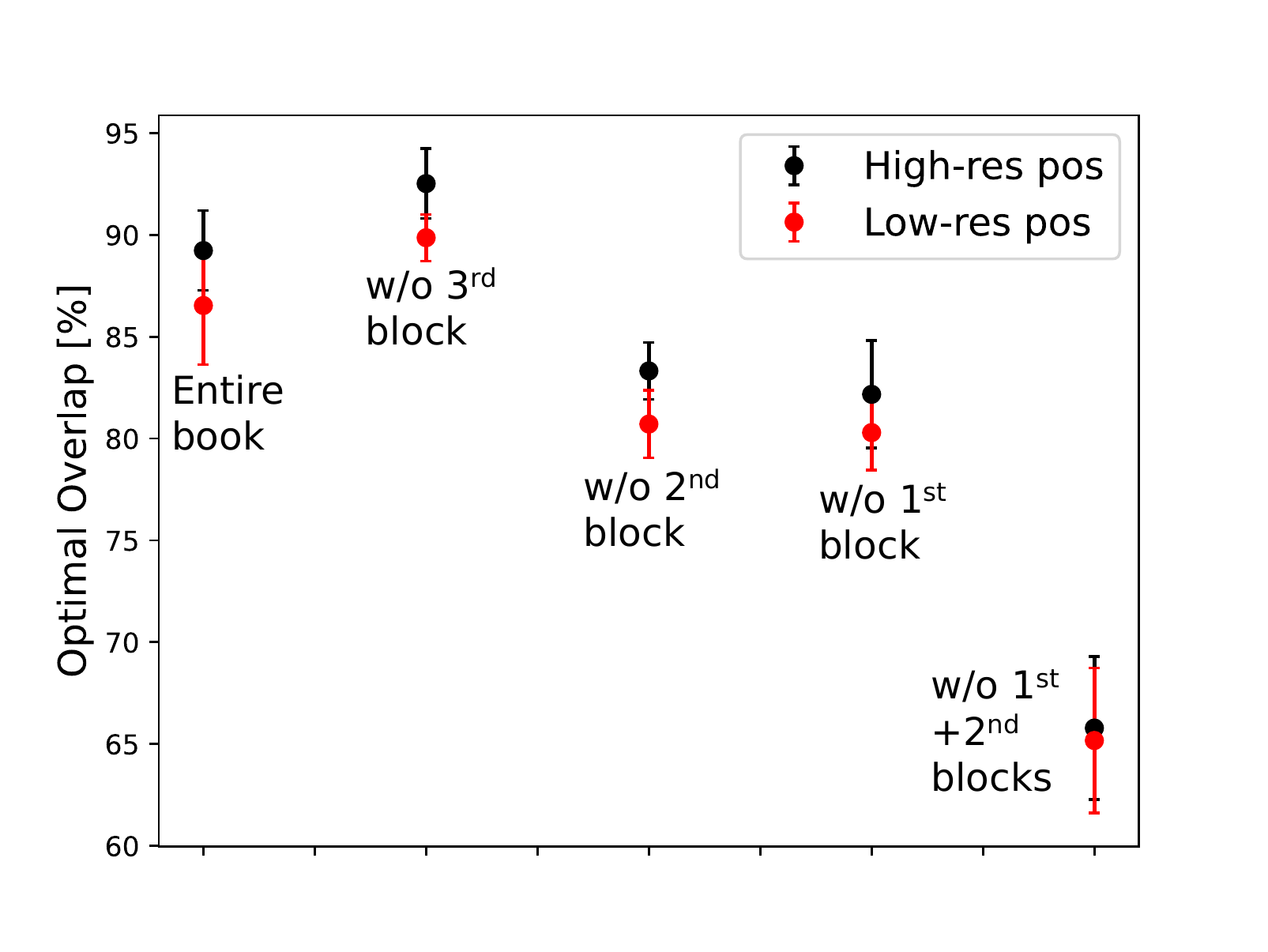}
  \caption{Optimal overlap of a cross-validated optimization analysis of the book of Exodus, after removing different P-associated blocks. \textbf{Black dots}: high-res POS. \color{red} \textbf{Red dots}\color{black}: low-res POS.}
     \label{Fig_Exodus_acc_blockRemoval}
\end{figure}

To examine this, we perform the following experiment: each time, we remove a different P-associated block (1st largest, 2nd largest, 3rd largest, and the 1st + 2nd largest) from the text and perform a cross-validated optimization analysis with low- and high-res POS (see \S\ref{methods_optimization}). We then compare the resulting optimal overlap values of each time. We plot the results of this experiment in Fig.~\ref{Fig_Exodus_acc_blockRemoval}.

We find that, as expected, the optimal overlap drops as a function of the size of the removed P-associated text. Interestingly, the optimal overlap \textit{increases} when a smaller block is removed. Additionally, unlike in the case of Genesis, the low-res POS representation doesn't lead to increased optimal overlap relative to the high-res POS representation. This suggests that: (\textbf{1}) the fluctuation of the optimal overlap indicates that our pipeline is sensitive to some semantic field associated with the two large P blocks rather than to a global stylistic difference between P/nonP. (\textbf{2}) In cases of more sporadically-distributed texts that are stylistically different from the text in which they are embedded -- one representation of the text is not systematically preferable to others.

\section{Conclusions} \label{conclusions}

We examined the hypothetical distinction between texts of priestly (P) and non-priestly (nonP) origin in the books of Genesis and Exodus, which we explored with a novel unsupervised pipeline for text stylometry. We sought a combination of a running-window width (i.e., the number of consecutive verses to consider as a single unit of text) and $n$-gram size of lexemes, low- or high-resolution phrase-dependent parts-of-speech that optimized the overlap between the unsupervised and hypothesized partitions. We established the statistical significance of our results using a cyclic-shift test, which we show to be more adequate for text stylometry problems than a naive permutation test. Finally, we extracted $n$-grams that contribute the most to the classification, their respective proportions, statistical robustness, and correlation to other features. We achieve optimal, statistically significant overlap values of 73\% and 90\% for the books of Genesis and Exodus, respectively. 

We find the discrepancy in optimal overlap values between the two books to stem from two factors: (\textbf{1}) A more sporadic distribution of P texts in Genesis, as opposed to a more formulaic one in Exodus. (\textbf{2}) The sensitivity of our pipeline to a distinct semantic field manifested in two large P blocks in Exodus, comprising the majority of the P-associated text therein. 

Through complementary exegetical and statistical analyses, we show that our methodology differentiates the unique generic style of the Priestly source, characterized by lawgiving, cult instructions, and streamlining a continuous chronological sequence of the story through third-person narration. This observation corroborates and hones the stance of most biblical scholars.

\section{Limitations} \label{limitations}

\begin{trivlist}
\item[~$\bullet$] The interdisciplinary element -- at the heart of this work -- mandates that our results be interpretable and relevant to scholars from the opposite side of the methodological divide (i.e., biblical scholars). This, in turn, introduces constraints to our framework -- the foremost is choosing appropriate text-embedding techniques. As discussed in \S\ref{methods_embedding} and \S\ref{methods_interpretability}, the ability to extract specific lexical features (i.e., unique $n$-grams) that are important to the classification, to quantify them, and subject them to complementary philological analysis (see Appendix \ref{app_disc}) -- requires that they be interpretable. This constraint limits the ability to implement state-of-the-art language-model-based embeddings without devising the required framework for their interpretation. Consequently, using traditional embeddings -- which encode mostly explicit lexical features (e.g., see \S\ref{methods_embedding}) -- limits the complexity of the analyzed textual phenomena and is therefore agnostic of potential signal that is manifested in more complex features.

\item[~$\bullet$] In text stylometry questions, especially those related to ancient texts, it is often problematic (and even impossible) to rely on a benchmark training set with which supervised statistical learning can take place. This, in turn, means that supervised learning in such tasks must be implemented with extreme caution so as not to introduce a bias into a supposedly-unbiased analysis. Therefore, implementing supervised learning techniques for such tasks requires a complementary framework that could overcome such potential biases. In light of this, our analysis involves predominantly unsupervised exploration of the text, given different parameterizations.

\item[~$\bullet$] Our ability to draw insight from exploring the stylistic differences between the hypothesized distinct texts relies heavily on observing significant overlap between the hypothesized and unsupervised partitions. Without it, the ability to discern the similarity between the results of our pipeline is greatly obscured, as the pipeline remains essentially agnostic of the hypothesized partition. Such a scenario either deems the parameterization irrelevant to the hypothesized partition or disproves the hypothesized partition. Breaking the degeneracy between these two possibilities may entail considerable additional analysis.
\end{trivlist}

\section{Acknowledgements} \label{acknowledgements}

We thank Dr. Rotem Dror for her kind assistance and contribution to the editing process and Ziv Ben-Aharon for providing technical guidance in operating the HUJI cluster.

\FloatBarrier
\bibliography{anthology,custom}
\bibliographystyle{acl_natbib}

\newpage
\appendix
\part*{Appendices}

\section{Estimating Scholarly Consensus of P-associated Texts in Genesis and Exodus} \label{app_consensus}

To quantify the consensus amongst biblical scholars concerning the distinction between P/nonP texts in Genesis and Exodus, we consider two sets of labelings of P/nonP: the first is that provided by biblical scholars, and the other is similar labeling used in the work of \citet{dershowitz2015computerized}, who also apply computational methods in an attempt to detect a meaningful dichotomy between P and nonP texts as well, albeit under a different paradigm. In their work, Dershowitz et al.\@ consider P/nonP distinctions of three independent biblical scholars and compile a single ``consensus'' labeling of verses over the affiliation of which all three scholars agree (1291 verses in Genesis and 1057 verses in Exodus), except for explicitly incriminating P texts (such as genealogical trees that are strongly affiliated with P \citep[e.g.,][]{jonker2012reading}). We find a 96.5\% and 97.3\% agreement between the two labelings for the books of Genesis and Exodus, respectively. 

\section{Cross-validated Overlap Optimization}

The cross-validated overlap optimization is performed as follows:

\begin{itemize}
    \item For lexemes/POS, we consider a range of $n$-gram sizes and a range of running-window widths, through combinations of which we optimize the classification overlap. These ranges are determined empirically by observing the overlap values decrease monotonically as the $n$-gram sizes/running-window widths become too large (see Figs. \ref{Fig_Optim_SingleFeature_Exodus}, \ref{Fig_Optim_SingleFeature_Genesis}). This produces a 2D matrix, where the $(i,j)${th} entry is the resulting overlap value achieved by the $k$-means given the combination of the $i${th} running-window width and the $j${th} $n$-gram size.
    \item We cross-validate the grid-search process by performing a series of simulations, where in each simulation, we generate a random subset of 250 verses from the given book, containing at least 50 verses of each class (according to the scholarly labeling).  Each such simulation produces a 2D overlap matrix, as mentioned above, for the given subset of verses.
    \item Finally, we average the 2D overlap matrices of all simulations, and for each entry, we calculate its standard deviation across all simulations, producing a 2D standard deviation matrix. After normalizing the average overlap matrix by the standard deviation matrix, we choose the optimal combination that produces the classification with the optimal overlap.
    \item We perform this analysis on both lexemes, low- and high-res POS, yielding three averaged overlap matrices, from which optimal parameter combinations and their respective standard deviations (calculated across the cross-validation simulations) can be identified for each feature.
\end{itemize}

\label{app_optimization}

\section{Feature Importance Analysis}

We consider the two optimal overlap clusters of verses as the algorithm assigned them. We calculate a difference matrix $D$, where from every verse in cluster 1, we subtract every verse in cluster 2, receiving a matrix $D \in \mathbb{R}^{|S_1|\cdot |S_2| \times d}$, where $|S_1|$, $|S_2|$ are the number of verses assigned to cluster 1 and 2, respectively, and $d$ is the dimension of the embedding (i.e., number of unique $n$-grams in the text). Note that tf-idf embedded texts are non-negative, such that the difference matrix $D$ has positive values for features with a high tf-idf score in cluster 1 and negative values for features with a high tf-idf score in cluster 2. As mentioned in \S\ref{methods_interpretability}, the first principal axis of $D$ is equivalent to the difference between the two centroids produced by the $k$-means. Similarly, this difference vector is a linear combination of all the features (i.e., $n$-grams), where a numerical value gives the importance of each feature called ``loading'', ranging from negative to positive values and their importance to the given principal axes is given by their absolute value. Due to the nature of the difference matrix $D$ (see \S\ref{methods_interpretability}) -- the sign of the loading indicates in which cluster the feature is important. Thus we can assign distinguishing features to the specific cluster of which they, or a combination thereof with other features, are characteristic.

Finally, we seek to determine the stability of the importance of features across multiple sub-samples of each book. We perform this computation as follows: given a parameters combination, we perform all the steps listed above for 100 simulations, where in each simulation, we randomly sample a sub-sample of 250 verses and extract the importance loadings of the features. We then average all simulation-wise loadings and derive the variance thereof to receive a cross-validated vector of feature importance loadings and their respective uncertainties. These are plotted as the error bars in Figs. \ref{Fig_PnonP_Genesis}, and similarly in the figures in Appendix \ref{app_features}

\label{app_features}

\section{Results} \label{app_results}

\subsection{Optimization Results: Genesis} \label{app_Results_SingleOptim_Genesis}

\begin{figure*}[t]
\begin{subfigure}[a]{\textwidth}
\centering
\includegraphics[scale = 0.45]{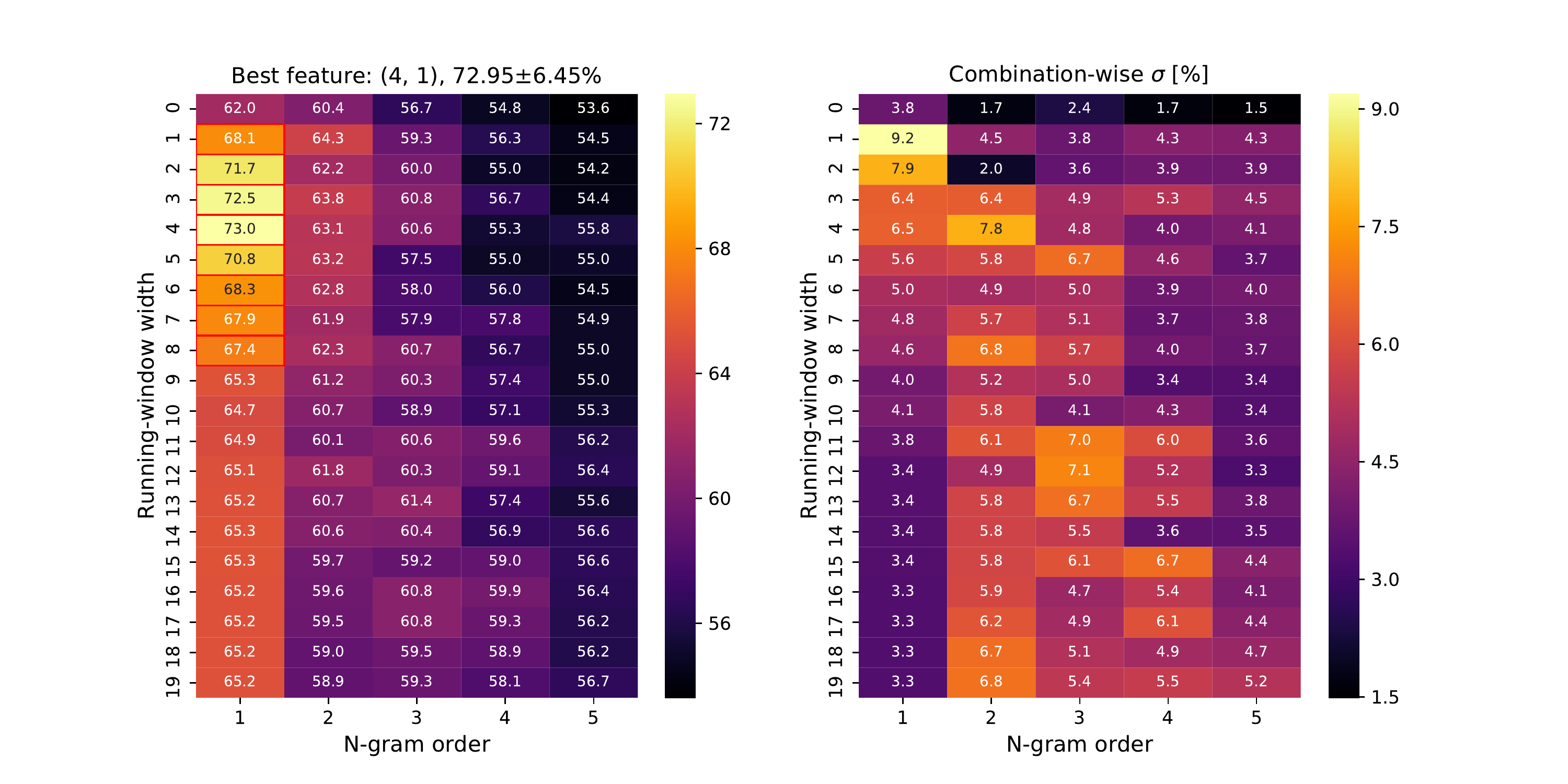}
\subcaption{Genesis cross-validated optimization analysis results: \textbf{lexemes}.}
\label{Optim_Genesis_SingleFeature_Words}
\end{subfigure} \\
\begin{subfigure}[b]{\textwidth}
\centering
\includegraphics[scale = 0.45]{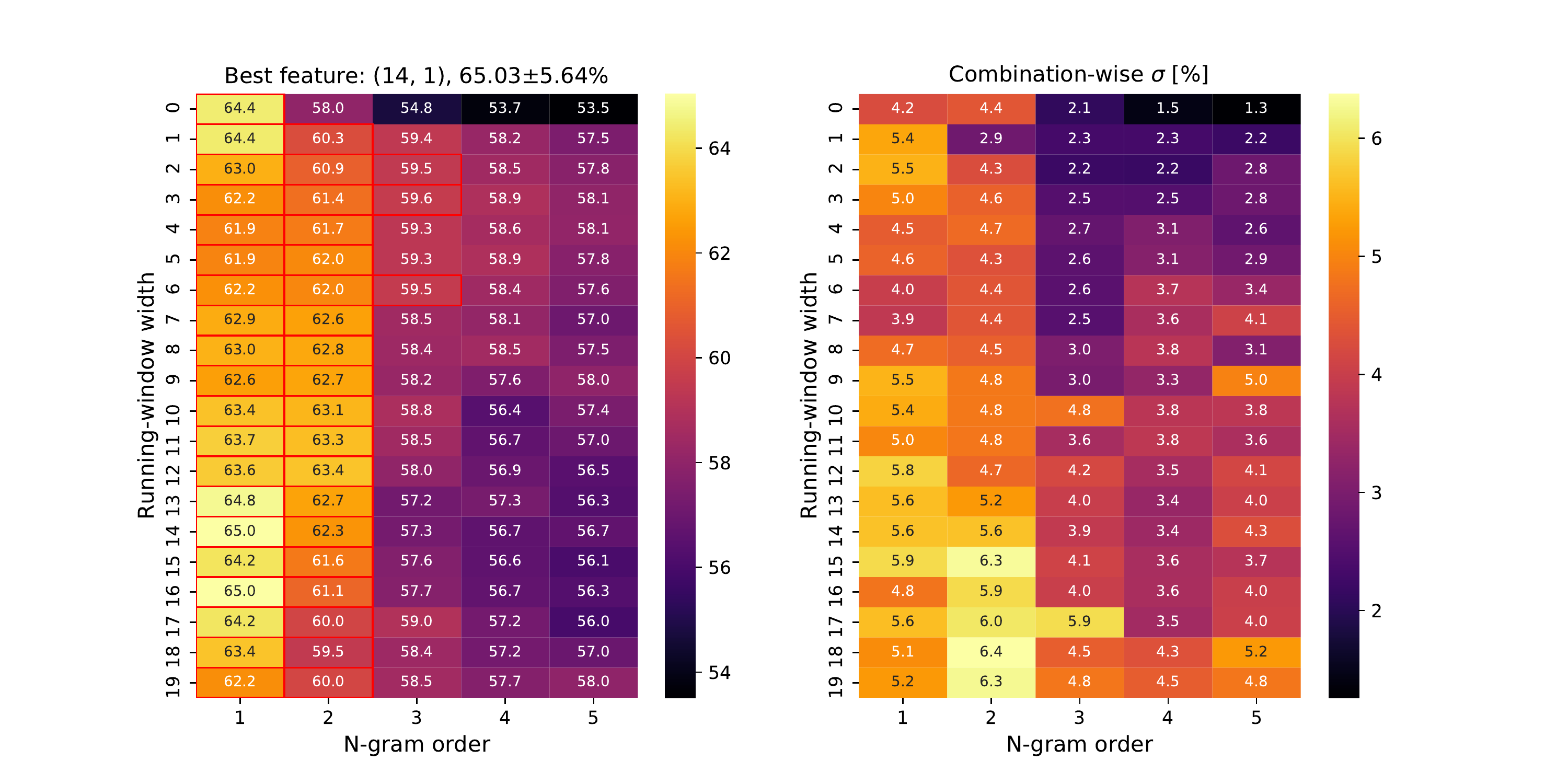}
\subcaption{Genesis cross-validated optimization analysis results: \textbf{high-res POS}.}
\label{Optim_Genesis_SingleFeature_TOTHT}
\end{subfigure}

\begin{subfigure}[c]{\textwidth}
\centering
\includegraphics[scale = 0.45]{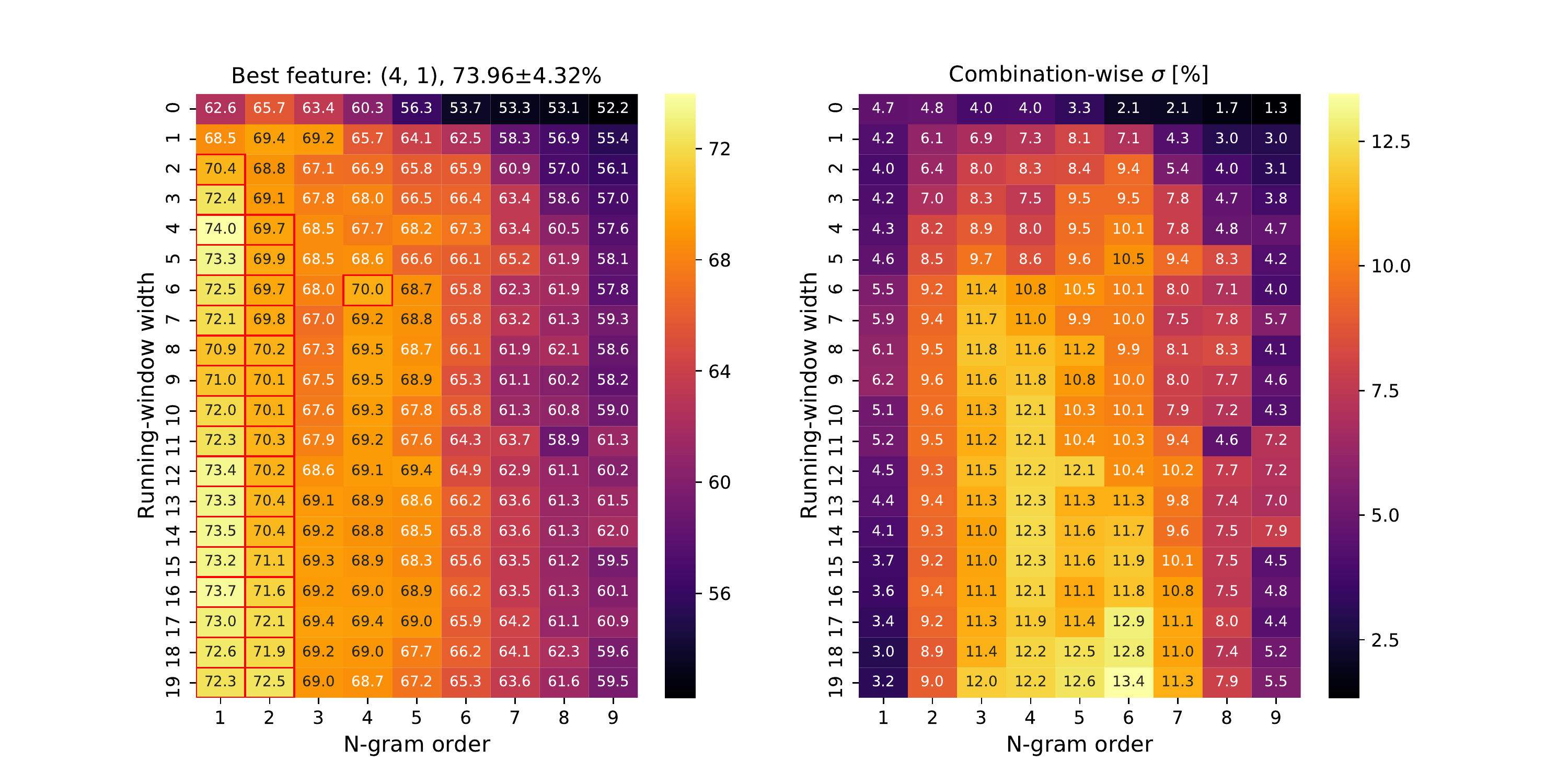}
\subcaption{Genesis cross-validated optimization analysis results: \textbf{low-res POS}.}
\label{Optim_Genesis_SingleFeature_BHSA}
\end{subfigure}
\caption{Cross-validated optimization analysis results for the book of Genesis, using all three representations: lexemes (\textbf{a}), high- and low-res POS (\textbf{b}, \textbf{c}), respectively, probing a running-window width range of 1--10 and $n$-gram size range of 1--10 over 20 simulations. \textbf{Left Panels}: Averaged combination-wise overlaps. \textbf{Right Panels}: Combination-wise standard-deviation ($\sigma$) of the overlaps generated with the cross-validation process. For each case, the best-fit feature (running-window width, $n$-gram size) and its respective $\sigma$ appear at the top of each left panel. Overlaps of combinations within the $1\sigma$ range of the best fit are marked with red cells.}
\label{Fig_Optim_SingleFeature_Genesis}
\end{figure*}

For each representation, we achieve the following optimal overlap values (see Fig.\@ \ref{Fig_Optim_SingleFeature_Exodus}): 72.95$\pm$6.45\% for lexemes, 65.03$\pm$5.64\% for high-res POS, and 73.96$\pm$2.91\% for low-res POS. We observe the following:

\begin{itemize}
    \item Optimal overlap values achieved for lexemes and high-res POS are consistent to within a 1$\sigma$, whereas the optimal overlap achieved for low-res POS is higher by $\sim2\sigma$.

    \item We find a less consistent and considerably larger spread of optimal parameter combinations for the low- and high-POS-wise embeddings, as opposed to Exodus. While no clear pattern of optimal parameter combinations is observed in any feature, small $n$-gram sizes are also preferred here. For lexemes, a well-defined range of running-window widths of unigrams is observed to yield optimal overlaps.

    \item Unlike in the case of Exodus, parameter combinations yielding optimal overlap values of each feature (marked with red cells in Fig.\@ \ref{Fig_Optim_SingleFeature_Genesis}) do not exhibit higher consistency across the cross-validation simulations than combinations that yield smaller overlap (i.e., small cross-validation variance, see the right panels in Fig.\@ \ref{Fig_Optim_SingleFeature_Genesis}), except for the low-res POS feature.

\end{itemize}

\subsection{Optimization Results: Exodus} \label{app_Results_SingleOptim_Exodus}

\begin{figure*}[t]
\begin{subfigure}[a]{\textwidth}
\centering
\includegraphics[scale = 0.45]{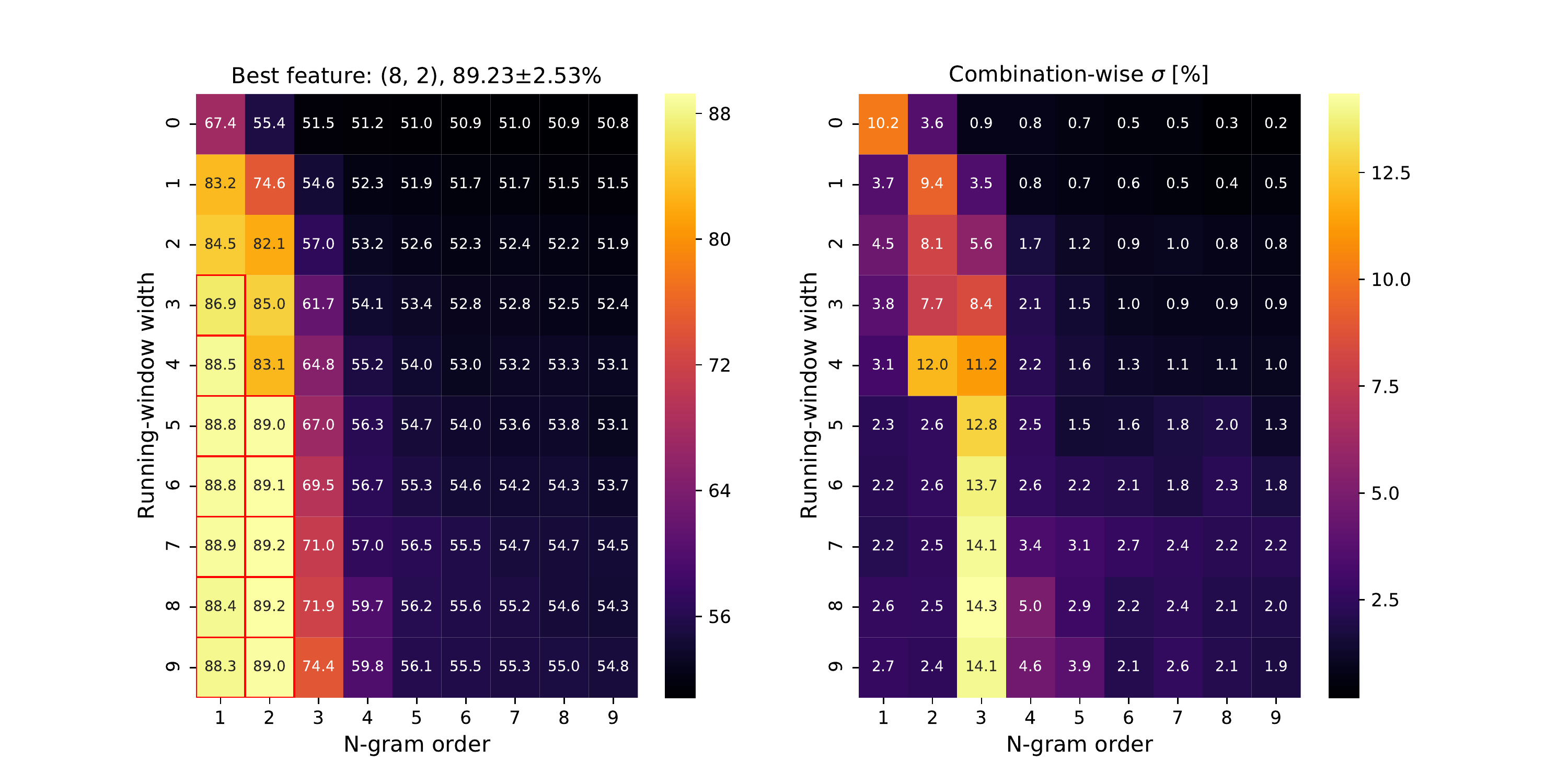}
\subcaption{Exodus cross-validated optimization analysis results: \textbf{lexemes}.}
\label{Optim_Exodus_SingleFeature_Words}
\end{subfigure} \\
\begin{subfigure}[b]{\textwidth}
\centering
\includegraphics[scale = 0.45]{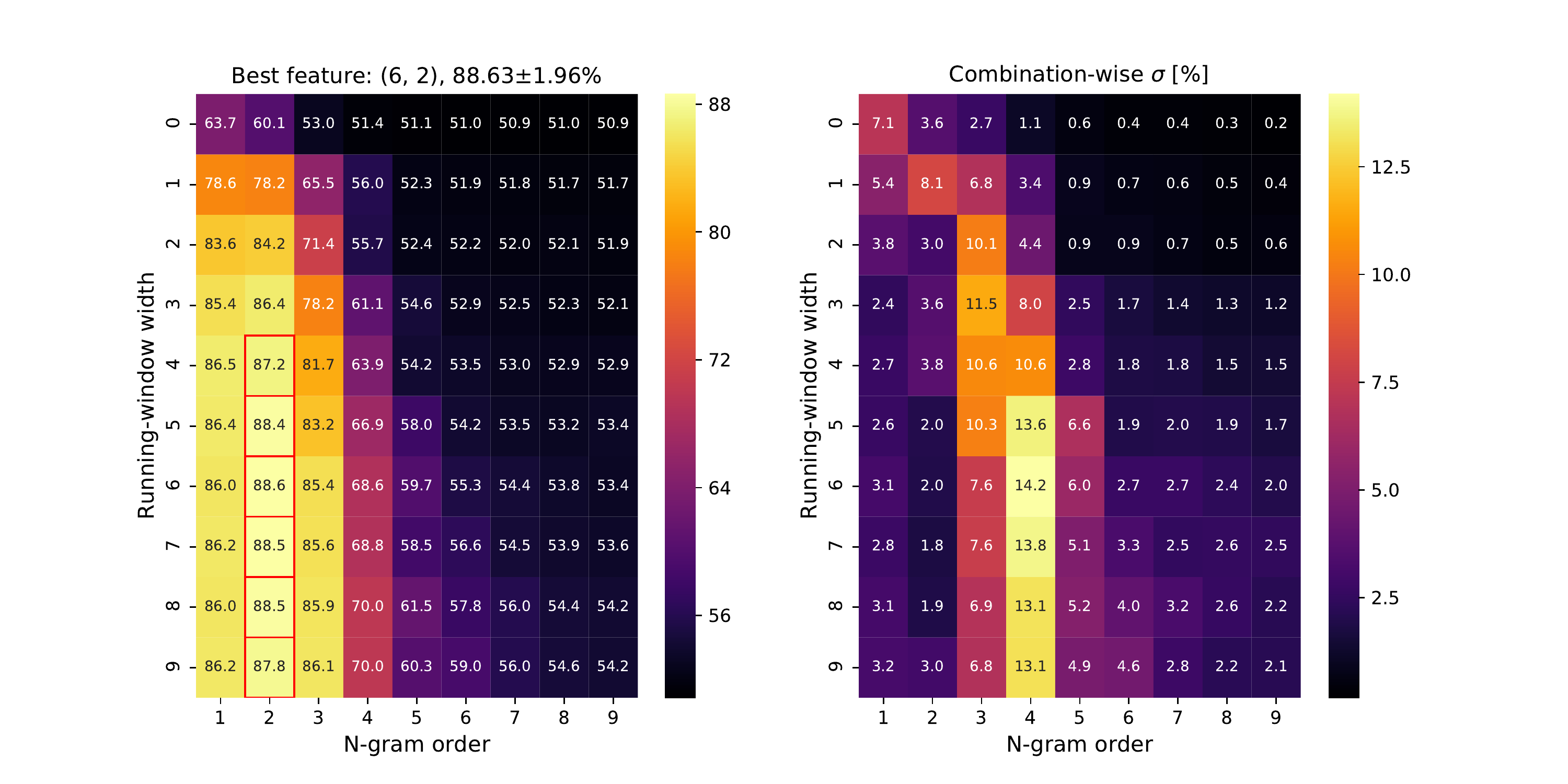}
\subcaption{Exodus cross-validated optimization analysis results: \textbf{high-res POS}.}
\label{Optim_Exodus_SingleFeature_TOTHT}
\end{subfigure}

\begin{subfigure}[c]{\textwidth}
\centering
\includegraphics[scale = 0.45]{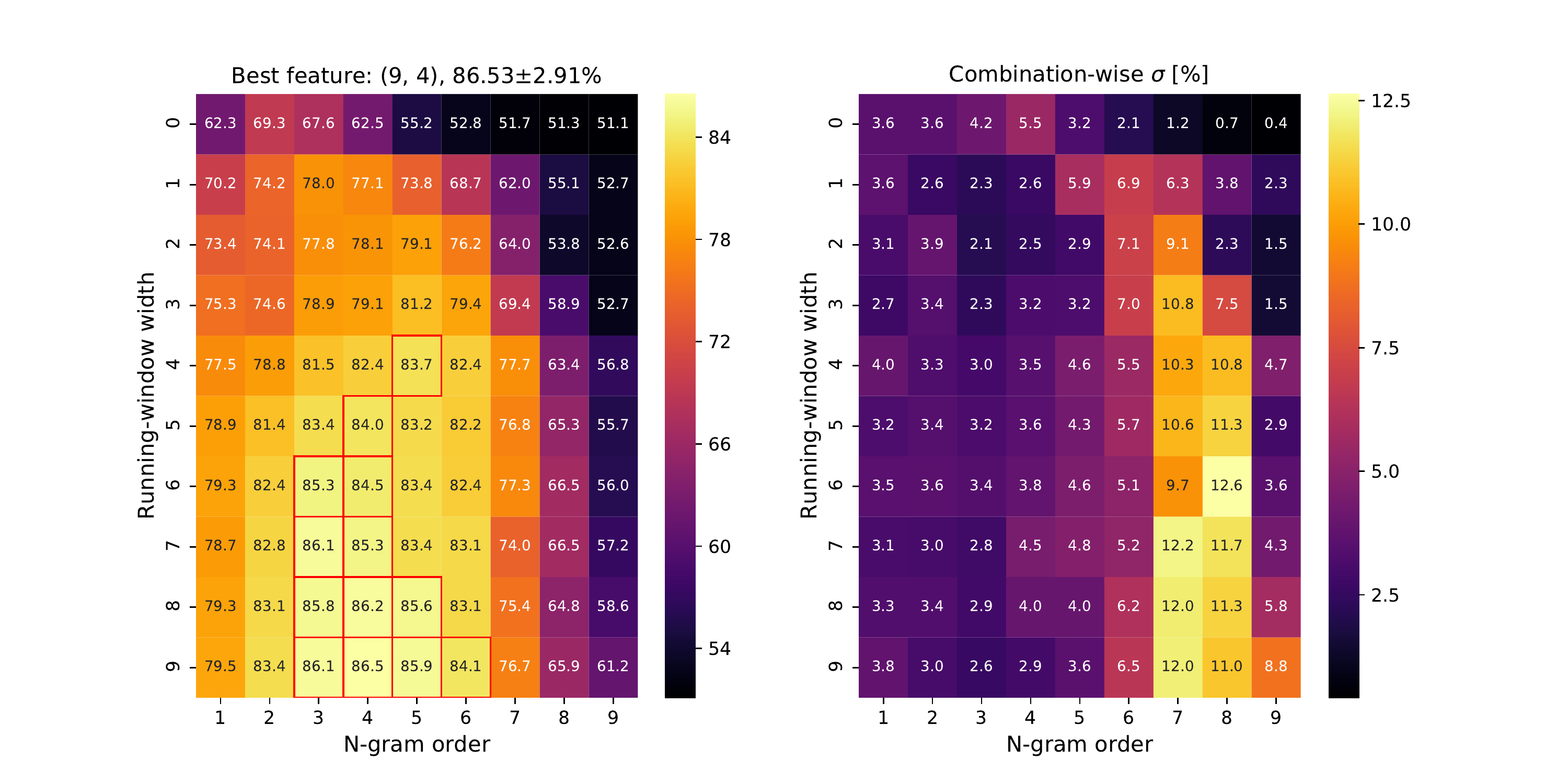}
\subcaption{Exodus cross-validated optimization analysis results: \textbf{low-res POS}.}
\label{Optim_Exodus_SingleFeature_BHSA}
\end{subfigure}
\caption{Cross-validated optimization analysis results for the book of Exodus, similar to Fig.\@ \ref{Fig_Optim_SingleFeature_Genesis}.}
\label{Fig_Optim_SingleFeature_Exodus}
\end{figure*}

For each feature, we achieve the following optimal overlaps (see Fig.\@ \ref{Fig_Optim_SingleFeature_Exodus}): 89.23$\pm$2.53\% for lexemes, 88.63$\pm$1.96\% for high-res POS, and 86.53$\pm$2.91\% for low-res POS. We observe the following:

\begin{itemize}
    \item For all three representations, optimal overlap values are consistent to within 1$\sigma$.

    \item For all three representations, parameter combinations yielding optimal overlap values (marked with red cells in Fig.\@ \ref{Fig_Optim_SingleFeature_Exodus}) exhibit high consistency across the cross-validation simulations (i.e., slight cross-validation variance, see the right panels in Fig.\@ \ref{Fig_Optim_SingleFeature_Exodus}).
    
    \item For lexemes, we find that the range of optimal parameter combinations is concentrated within 1- and 2-grams and is relatively independent of running-window width (i.e., when some optimal running-window width is reached--the overlap values do not change dramatically as it increases).

    \item For the high-res POS, we find that the range of optimal combinations is restricted to 2-grams, but is also insensitive to the running-window width.

    \item For low-res POS, we find that larger $n$-gram sizes, and a wider range thereof, produce the optimal overlap values. Additionally, we observe a dependence between given $n$-gram sizes and running-window widths to reach a large overlap.
\end{itemize}

\subsection{Hypothesis Testing Results} \label{Results_SingleHypothesis}

\begin{figure*}[p] 
\begin{subfigure}{\textwidth}
\centering
\includegraphics[scale = 0.5]{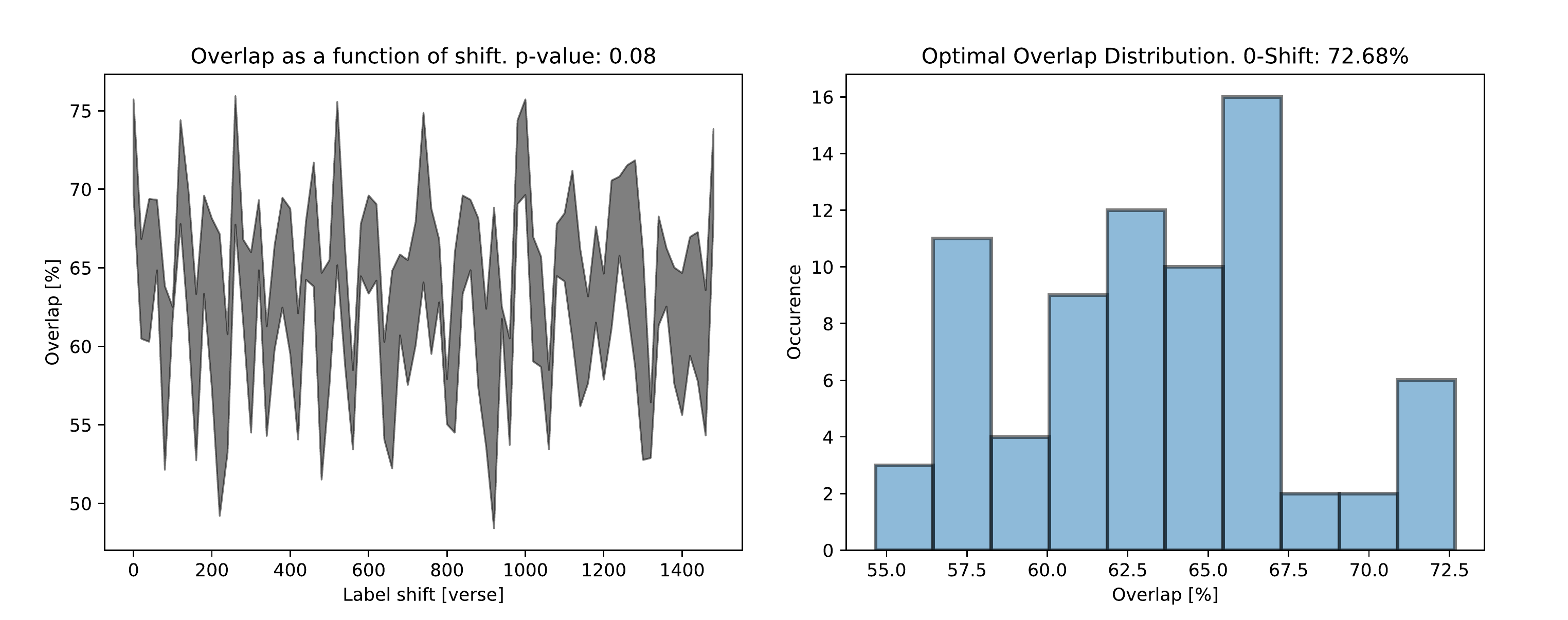}
\subcaption{Cross-validated hypothesis-testing results: book of Genesis (low-res POS)}
\end{subfigure} \\
\begin{subfigure}{\textwidth}
\centering
\includegraphics[scale = 0.5]{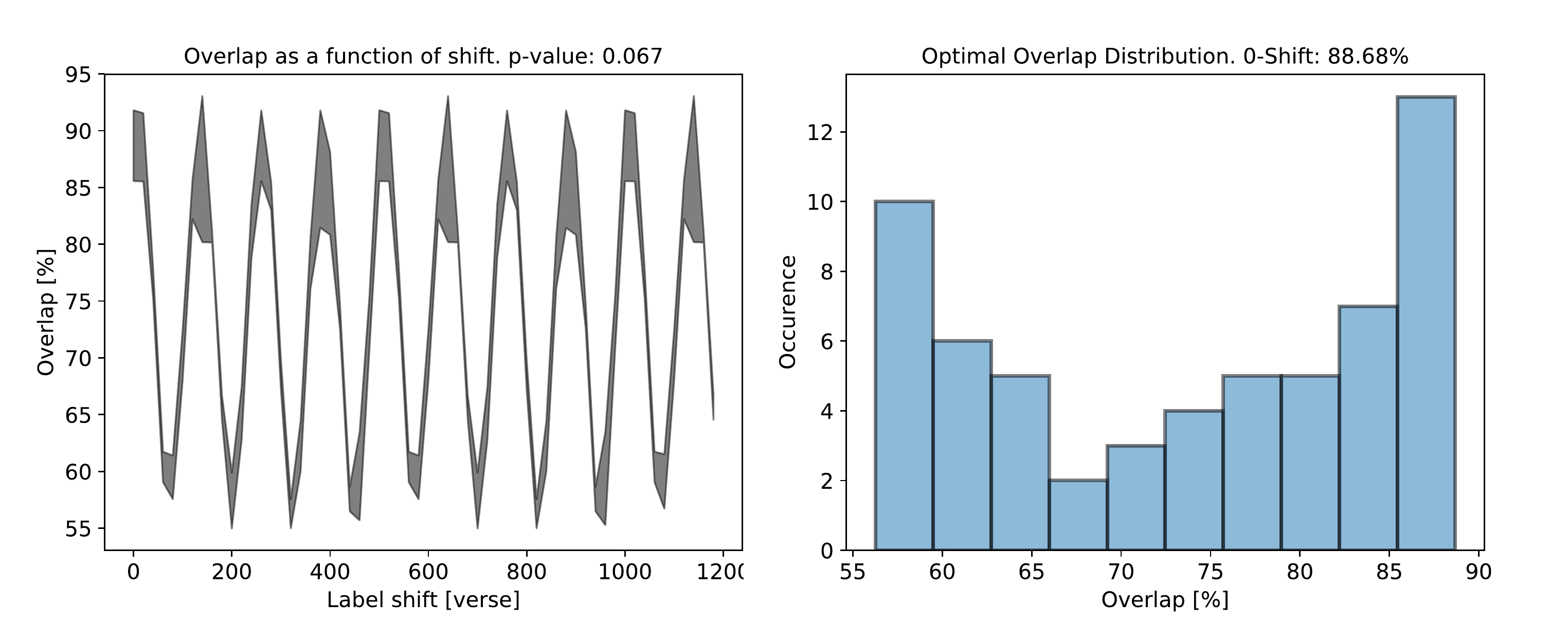}
\subcaption{Cross-validated hypothesis-testing results: book of Exodus (high-res POS)}
\end{subfigure}
\caption{Cross-validated hypothesis-testing results, performed as described in \S\ref{results}, for the books of Genesis (sub-figure \textbf{a}) and Exodus (sub-figure \textbf{b}), respectively. \textbf{Left Panels}: Optimal overlap as a function of cyclic label shift. The derived $p$-value is listed on top of each panel. \textbf{Right Panels}: Resulting null distributions of the optimal overlap values. The optimal overlap value of the 0-shift (i.e., scholarly labeling) is listed on each panel.}
\label{Fig_hypothesisTesting}
\end{figure*}

We present our results of the hypothesis testing through cyclic-shift, described in \S\ref{methods_hypothesisTesting}. Here, too, we perform a cross-validated test by performing five simulations -- each containing a randomly chosen sub-sample of 250 verses (with a mandatory minimum of 50 verses of each class), to which the cyclic-shift analysis is applied. We compute the optimal overlap for every shift and generate a shift-series of optimal overlap values (i.e., the null distribution). We then average across simulations. We then derive the $p$-value from the synthesized null distribution. The chosen ``real optimal overlap'', which we use to derive the $p$-value, is the average optimal overlap at a shift of 0 (i.e., original labeling) minus its standard deviation. For each book, we perform this analysis for the features yielding the optimal overlap; low-res POS for Genesis and high-res POS for Exodus. We plot our results for both books in Fig.\@ \ref{Fig_hypothesisTesting}.

The resulting $p$-values are 0.08 and 0.06 for the books of Genesis and Exodus, respectively.

\subsection{Feature Importance Analysis Results} \label{Results_Features}

\subsubsection{Feature Importance: Genesis}

\begin{figure*}[p] 
\centering
\includegraphics[scale = 0.5]{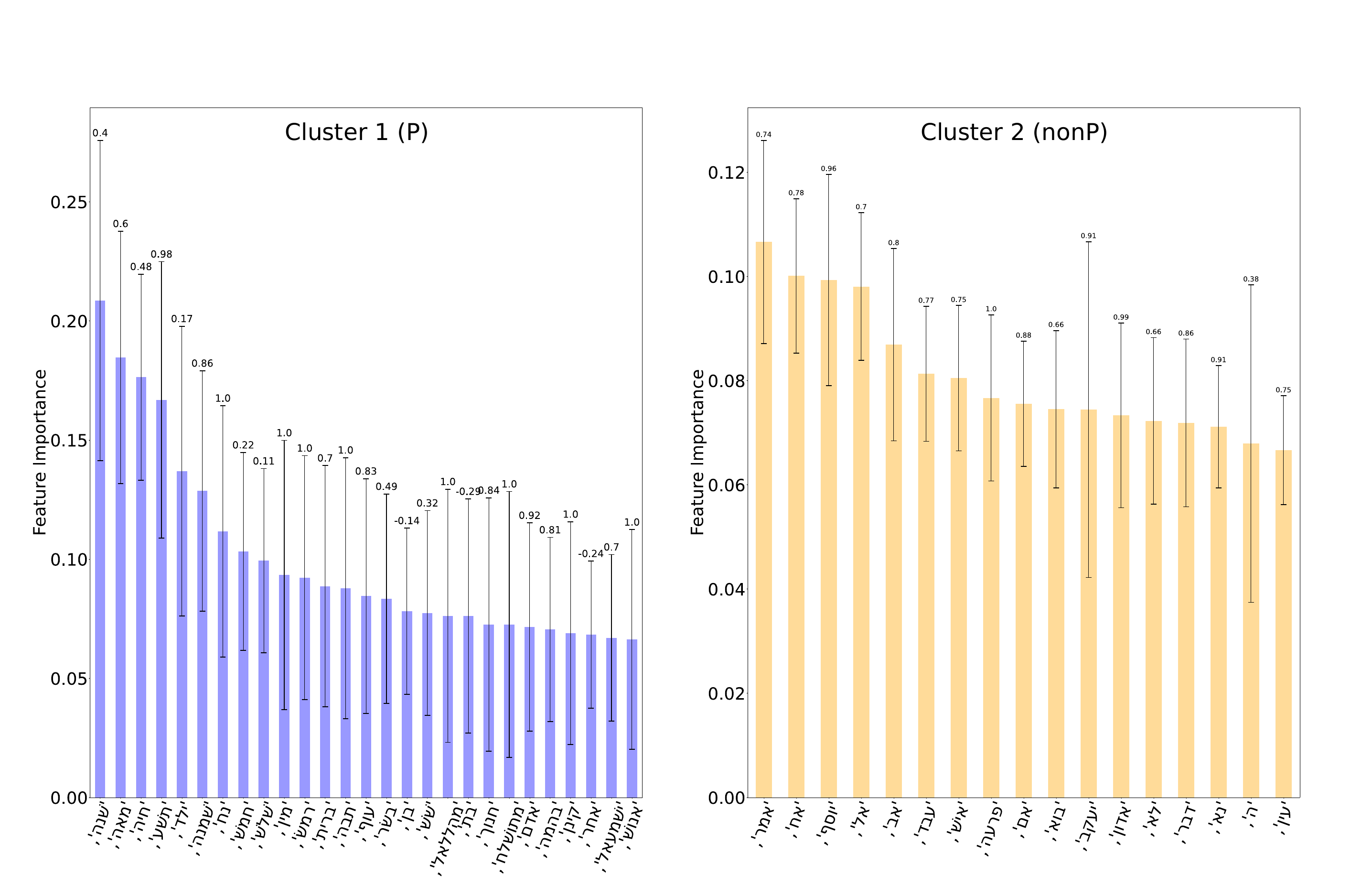}
\caption{Cross-validated feature importance analysis results for the book of Genesis: \textbf{lexemes} (running-window width of 4; $n$-gram size of 1). \textbf{Note}: displaying features carrying 75\% of the explained variance.}
\label{Fig_Genesis_FI_lexemes}
\end{figure*}

\begin{figure*}[p] 
\centering
\includegraphics[scale = 0.5]{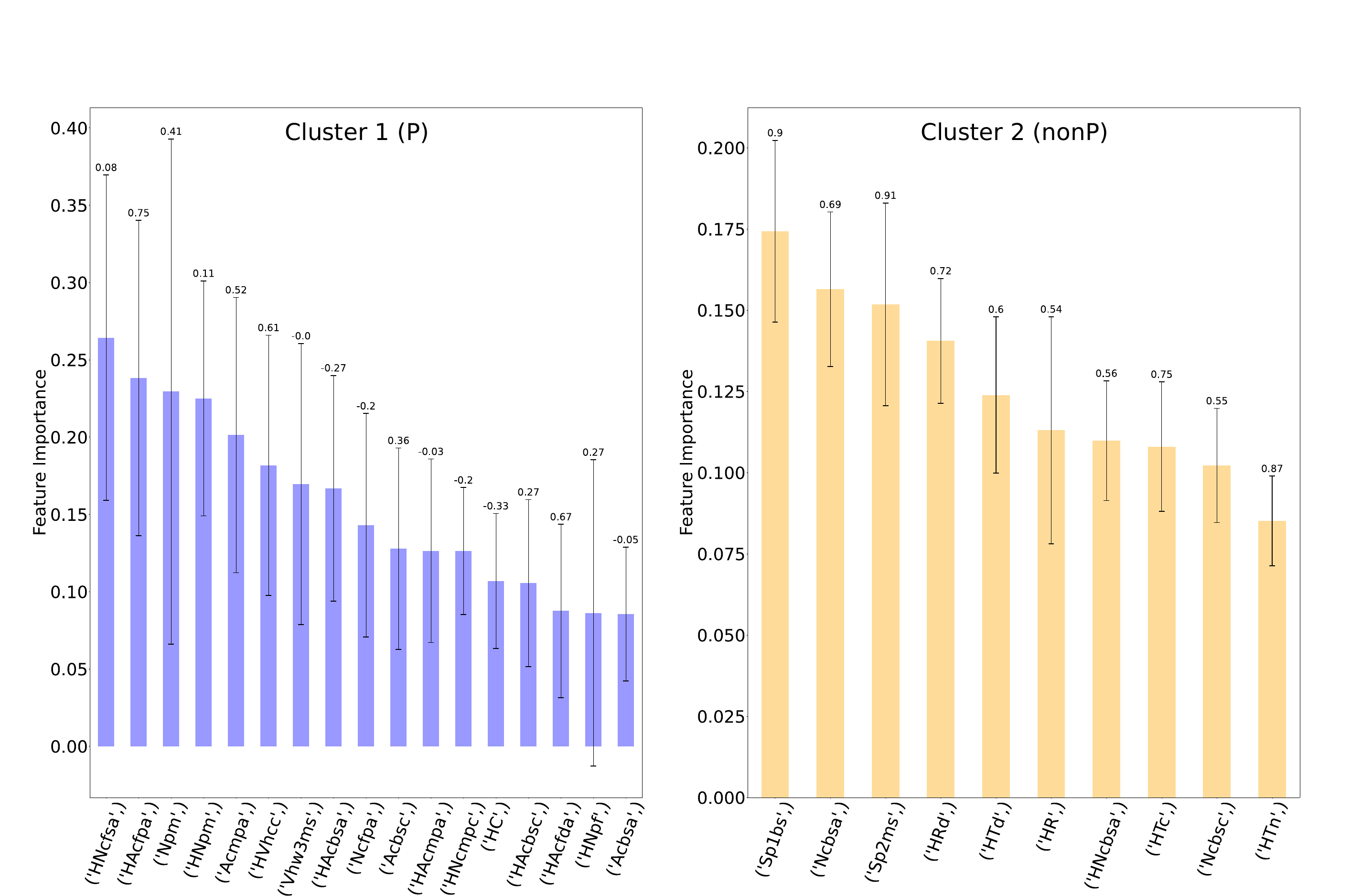}
\caption{Cross-validated feature importance analysis results for the book of Genesis: \textbf{high-res POS} (running-window width of 1; $n$-gram size of 1). \textbf{Note}: displaying features carrying 90\% of the explained variance.}
\label{Fig_Genesis_FI_highPOS}
\end{figure*}

\begin{figure*}[t!]
\centering
\includegraphics[scale = 0.5]{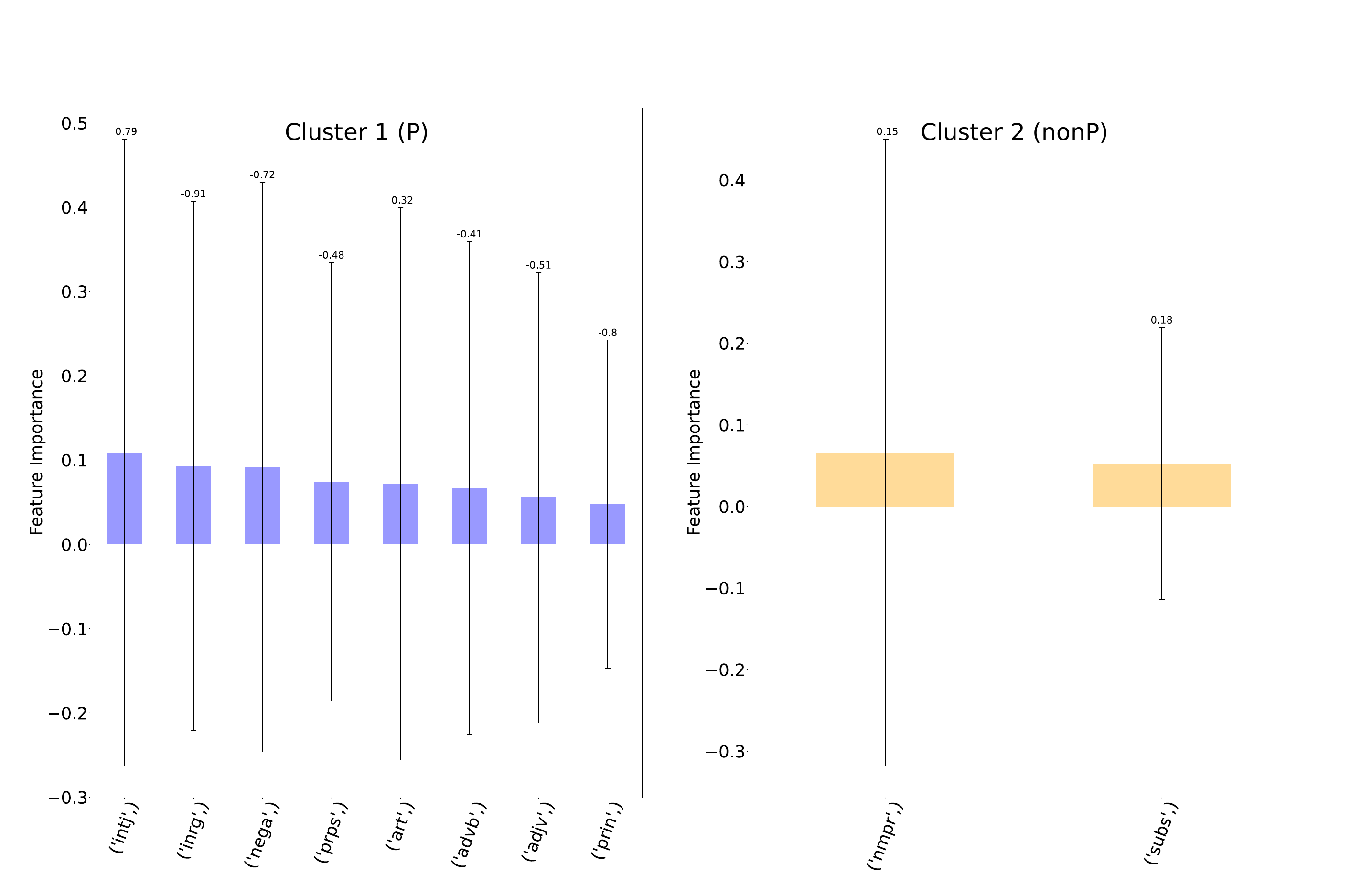}
\caption{Cross-validated feature importance analysis results for the book of Genesis: \textbf{low-res POS} (running-window width of 4; $n$-gram size of 1). \textbf{Note}: displaying features carrying 100\% of the explained variance.}
\label{Fig_Genesis_FI_lowPOS}
\end{figure*}

In Figs. \ref{Fig_Genesis_FI_lexemes}-\ref{Fig_Genesis_FI_lowPOS} we plot feature importance analysis results for the three representations of the book of Genesis.

\subsubsection{Feature Importance: Exodus}

\begin{figure*}[t!]
\centering
\includegraphics[scale = 0.55]{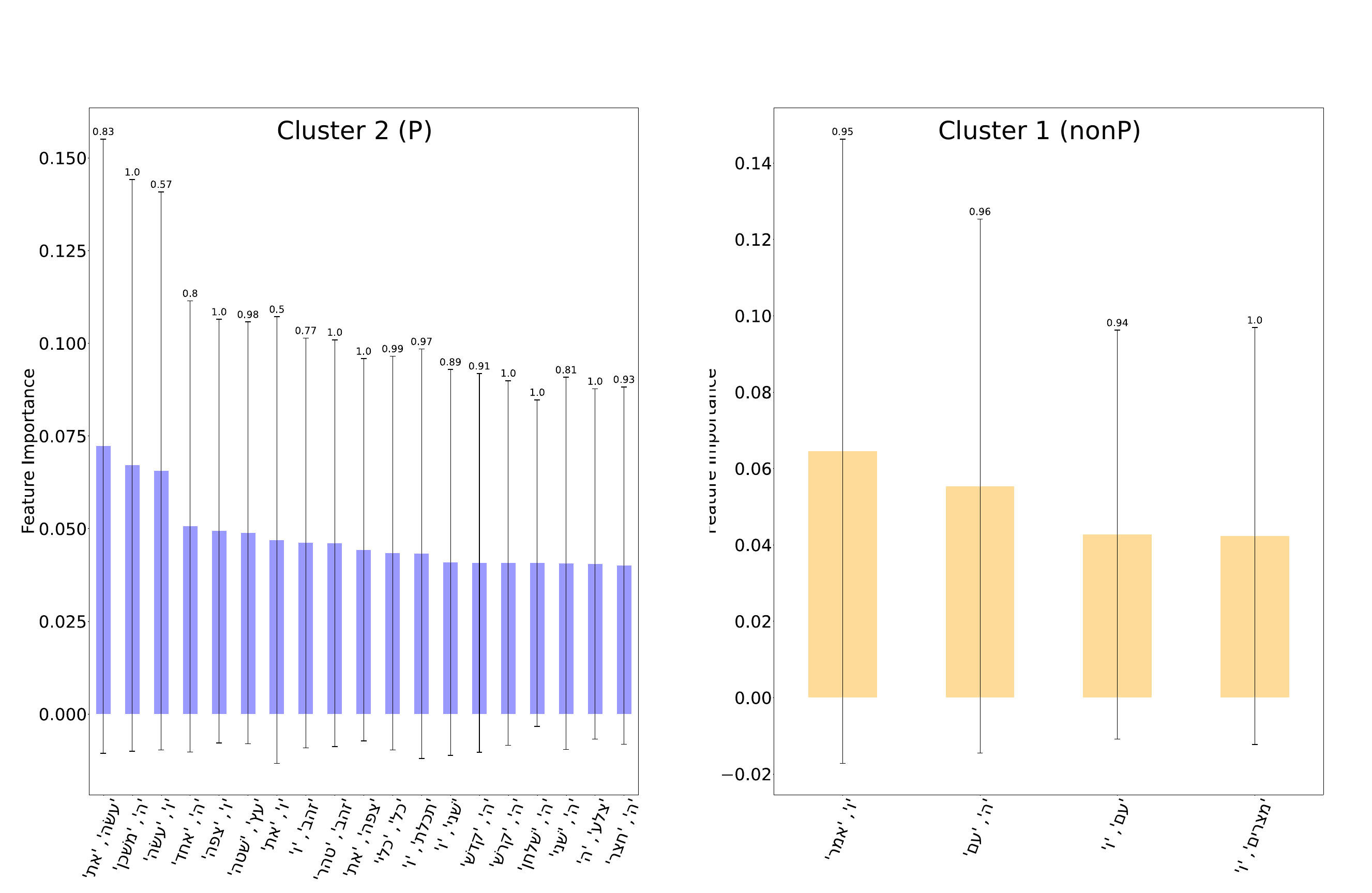}
\caption{Exodus cross-validated feature importance results: \textbf{lexemes} (running-window width of 5; $n$-gram size of 2).}
\label{Fig_Exodus_FI_lexemes}
\end{figure*}

\begin{figure*}[t!]
\centering
\includegraphics[scale = 0.55]{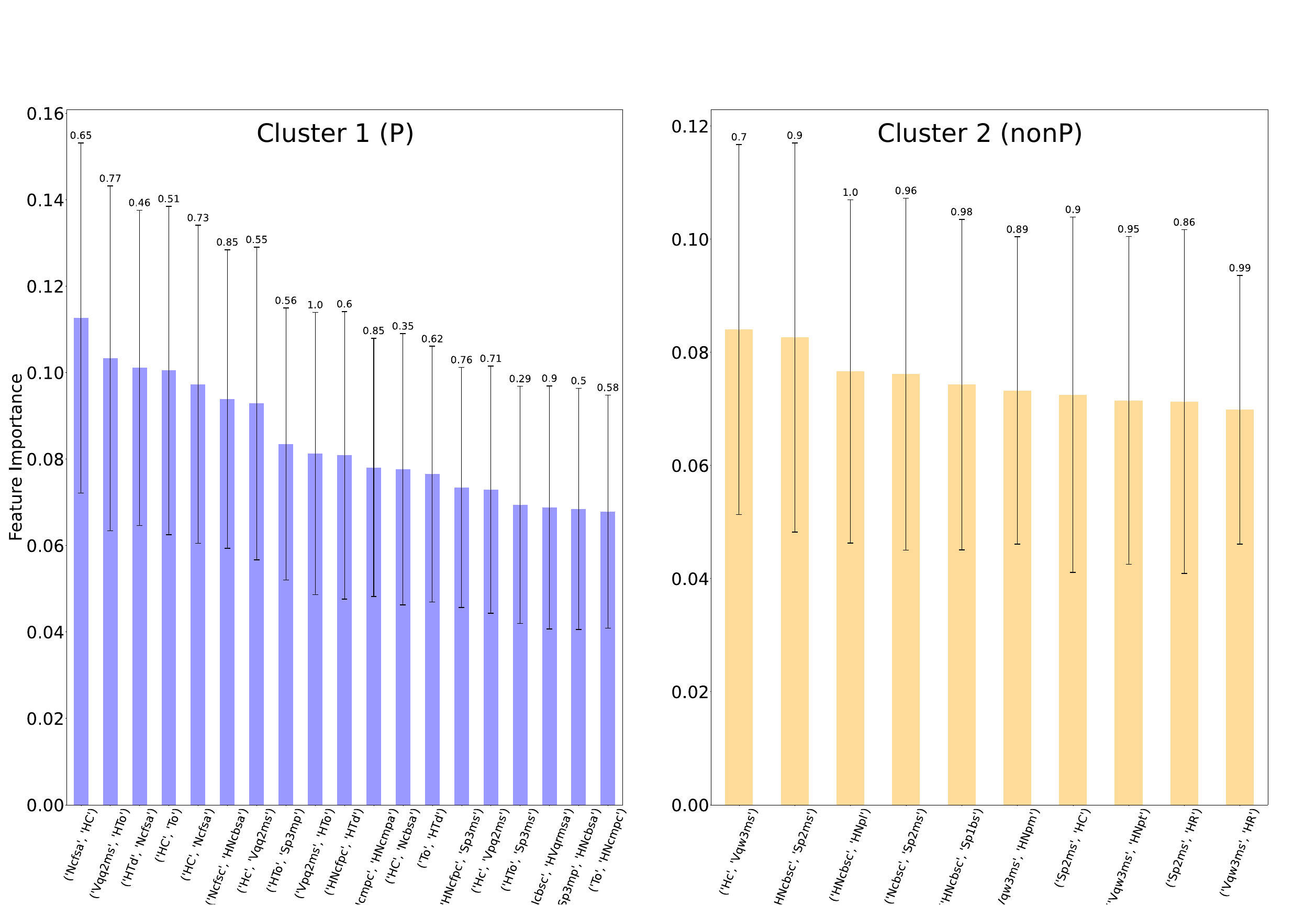}
\caption{Exodus cross-validated feature importance results: \textbf{high-res POS} (running-window width of 4; $n$-gram size of 2).}
\label{Fig_Exodus_FI_highPOS}
\end{figure*}

\begin{figure*}[t!]
\centering
\includegraphics[scale = 0.55]{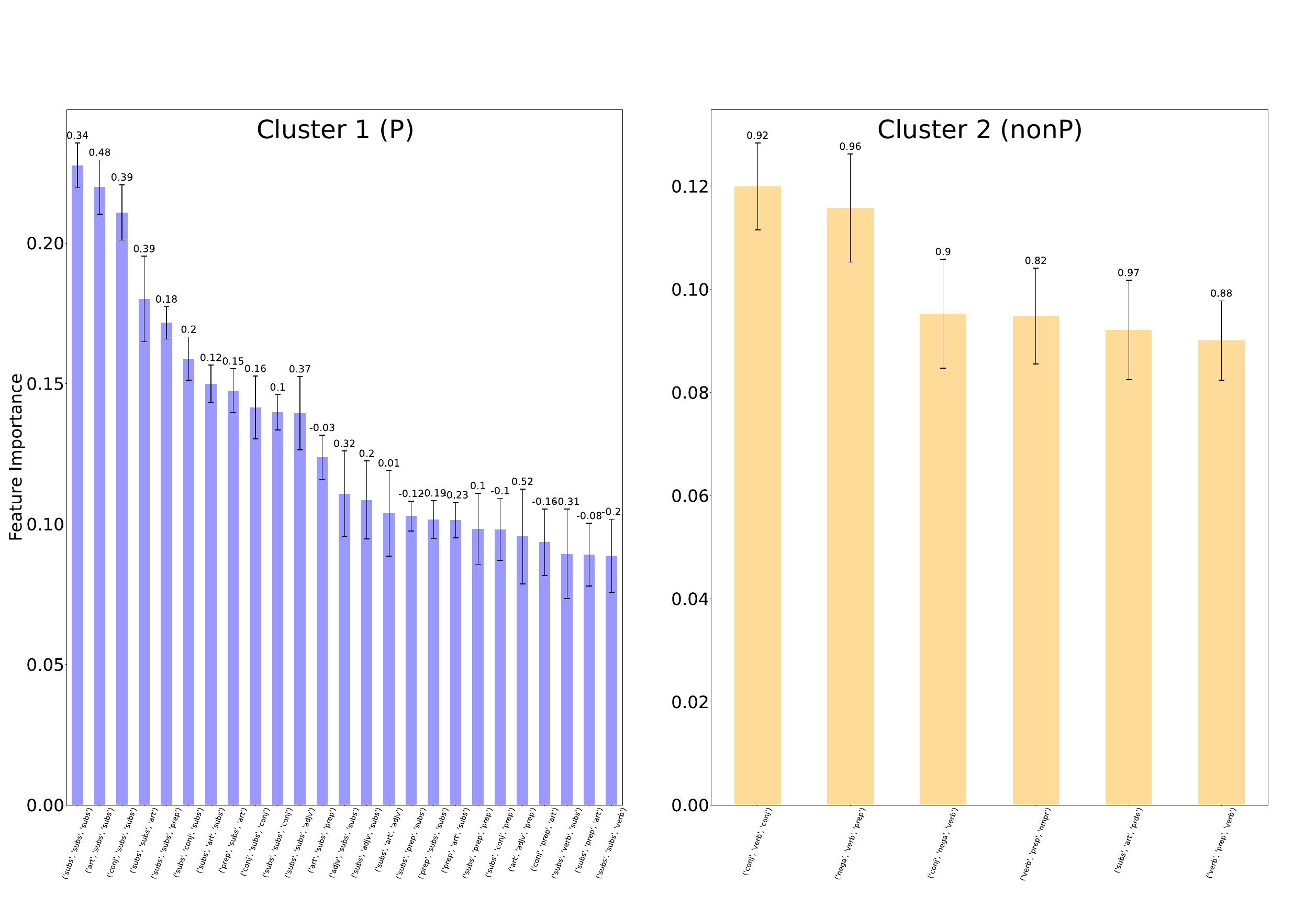}
\caption{Exodus cross-validated feature importance results: \textbf{low-res POS} (running-window width of 6; $n$-gram size of 3).}
\label{Fig_Exodus_FI_lowPOS}
\end{figure*}

In Figs. \ref{Fig_Exodus_FI_lexemes}-\ref{Fig_Exodus_FI_lowPOS} we plot feature importance analysis results for the three representations of the book of Exodus.

\section{Biblical-Exegetical Discussion} \label{app_disc}

Here we perform an exegetical analysis of our results for each book. All data to which this analysis was applied is available online\footnote{\url{https://github.com/YoffeG/PnonP}}. 

\subsection{Genesis} \label{app_disc_genesis}

\subsubsection{Semantics} \label{app_disc_genesis_semantics}

The extraction of the features of P for Genesis overlaps with the work of characterizing the priestly stratum \citep{holzinger1893einleitung}. Thus, the characteristic use of numbers in P (here, in descending order of importance, the algorithm considered the terms 100, 9, 8, 5, 3, 6, 4, 2, 7 as characteristic of P) appears mainly in the genealogies, e.g., Gen 5 and 11, but also in the use of ordinal numbers to give the months and in the definition of the dimensions of the tabernacle. The term "year" (\cjRL{+snh})  is used in both dates and P genealogies, and the term "day" (\cjRL{ywM}) demonstrates a similar calendrical concern. Furthermore, in the genealogies, we find the names of the patriarchs considered to belong to P (\cjRL{n.h}, \cjRL{mhll'l}, \cjRL{qynN}, \cjRL{.hnwK}, \cjRL{mtw+sl.h}, \cjRL{'nw+s}, \cjRL{'dM}, \cjRL{+st}, \cjRL{y+sm`'l}, \cjRL{lmK}, \cjRL{r`w}, \cjRL{plg}, \cjRL{srwg}). The term "son" (\cjRL{bN}) appears in genealogies but also in typical P expressions such as "son of X year" (\cjRL{bN +snh}) to indicate the age of someone, "sons of Israel" (\cjRL{bny y,sr'l}), etc. The root \cjRL{yld} (to beget or to give birth) is found in the genealogies in Gen 5; 11; Exod 6 but also in other P narratives of the patriarchs (Gen 16--17; 21; 25; 35; 36; 46; 48) which focus on affiliation.
The term "generation" (\cjRL{dwr}) is also recognized as typically P (Gen 6:9; 9:12; 17:7,9,12; etc. ), as well as the term "annals" (\cjRL{twldwt}), which serves to introduce a narrative section or a genealogy and. This term structures the narrative and genealogical sections in the book of Genesis (Gen 2:4; 5:1; 6:9; 10:1,32; 11:10,27; 25:12--13,19; 36:1,9; etc.).
The terms "fowl" (\cjRL{`wP}), "beast/flesh" (\cjRL{b,sr}), "creeping" (\cjRL{rm,s}), "swarming" (\cjRL{+sr.S}), "living being" (\cjRL{np+s .hyh}), "cattle" (\cjRL{bhmh}), "kind" (\cjRL{myN}) are found in the typically P expression "living creatures of every kind: cattle and creeping things and wild animals of the earth of every kind" (Gen 1:24; cf. Gen 1:25-26; 6:7,20; 7:14,23; etc.). These expressions are often associated with "multiplication" (\cjRL{rbh}), an essential theme for P that also appears in the blessings of P narratives as in Gen 17; 48; etc. The term "being" (\cjRL{np+s}) is also used in P texts to refer to a person, e.g., in Gen 12; 17; 36; 46. As for the term "all" (\cjRL{kl}), it is used overwhelmingly in both P and D texts.
In P texts (Gen 1:27; 5:2; 6:19), humanity (\cjRL{'dM}), in the image (\cjRL{.slM}) of God, is conceived in a dichotomy of "male" (\cjRL{zkr}) and "female" (\cjRL{nqbh}). The root \cjRL{zkr} in its second sense, that of "remembering", also plays a role in the P narratives (Gen 8:1; 9:15--16; 19:29; Exod 2:24; 6:5) when God remembers his covenant and intervenes to help humanity or the Israelites.
The covenant (\cjRL{bryt}; cf. also Gen 9; 17), the sign of which is the circumcision (\cjRL{mwl}) of the foreskin (\cjRL{`rlh}) is correctly characterized as P. According to P, God’s covenants are  linked to a promise of offspring (\cjRL{zr`}; cf. Gen 17; etc.) and valid forever (\cjRL{`wlM}; Gen 9; 17; 48:4; Exod 12:14; etc.). The term "seed/descendant" (\cjRL{zr`}) is also used by P in the creation narratives in Gen 1. The term "between" (\cjRL{byN}) is used several times to indicate the parties concerned by the covenant in Gen 9 and Gen 17. The term is also found frequently in Gen 1 in the creation story, where creation is the result of separation "between" (\cjRL{byN}) different elements -- presenting God as the creator is not typical of a national god whose role primarily guarantees protection, military success, and fertility. The transformation of the God of Israel into a creator God appears only in the exilic or postexilic texts. Thus, the root "to create" (\cjRL{br'}) is rightly associated with P (Gen 1:1-2:4; 5:1-2).
The use of divine names is particular in the priestly narratives. "God" (\cjRL{'lhyM}) is the term used in the origin stories (Gen 1--11), "El Shaddai" for the patriarchs (Gen 12--Exod 6), and finally, "YHWH" from Exodus 6,2-3 on. Here, the algorithm did understand a particular use by P of the term "God" (\cjRL{'lhyM}).
One of the differences with Holzinger’s list is the fact that the algorithm considers the terms Noah (\cjRL{n.h}), flood (\cjRL{mbwl}), and the ark (\cjRL{tbh}) as typically P. This is probably because the flood narrative is much more developed in P than in non-P or because the semantic environment is attached to other P expressions. Nevertheless, all three terms appear in non-P texts as well. The term "daughter" (\cjRL{bt}) should be considered P not because of its frequency, which is admittedly somewhat higher in the P narratives of Genesis, but probably because of its semantic environment. Thus, the term appears in the expression "sons and daughters" (\cjRL{bnwt wbnyM}), which is very frequently used in Gen 5; 11. The preposition "after" (\cjRL{'.hr})  appears in the expression "after you" (\cjRL{'.hryK}) in the promise to Abraham in Gen 17 or the expression "after his begetting" (\cjRL{hwlydw '.hry}) in the genealogies in Gen 5; 11. The appearance of the term "to die" (\cjRL{mwt}) as a characteristic of P is explained by its presence in the genealogies of Gen 5; 11 but also in the succession of each of the generations of the patriarchs. Finally, the terms "water" (\cjRL{myM}) and "heaven" (\cjRL{+smyM}) play a major role in the creation narrative P (Gen 1:1-2:4) and the flood narrative (Gen 6-9*). These two terms also appear in Exodus, where water is mentioned in the account of the duel of the magicians (Exod 7-9*), in the passage of the sea of reeds which is paralleled in the creation of Gen 1 (Exod 14), and as a means of purification during the building of the tabernacle (Exod 29-30; 40). This latter function of water probably builds its symbolism in the other narratives. The term firmament (\cjRL{rqy`}) is associated with heaven and appears only in the creation story P of Gen 1 but is of little significance elsewhere.

On the non-P side, terms like "Joseph", "pharaoh", and "Egypt" are non-P features since the story of Joseph is non-P. Similarly, the presence of "Jacob" is explained by an account of only a few verses for this story in P as opposed to several whole chapters for the non-P account of Jacob.
The terms "brother" (16$^{P}$/179$^{nonP}$), "father" (19$^{P}$/213$^{nonP}$), and "mother" (4$^{P}$/33$^{nonP}$) as non-P features can be understood by a greater emphasis on family in the original patriarchal accounts, whereas P emphasizes genealogy. The terms "master" (\cjRL{'dwN}), "slave/servant" (\cjRL{`bd}), and "boy/servant" (\cjRL{n`r}) reflect the hierarchical structures of the household of the wealthy landowners in the narratives of the patriarchs but are of no interest to the priestly editors. Similarly, non-P texts show more interest in livestock, with terms such as camel (\cjRL{gml}), donkey (\cjRL{.hmwr}), or small livestock (\cjRL{.s'N}) considered non-P features. The dialogues are more present in the non-P stories than in the P stories. Thus the terms that open the direct discourses "speak" (\cjRL{dbr}), "say" (\cjRL{'mr}), and "tell" (\cjRL{ngd}) are considered typical non-P terms as well as the set of Hebrew propositions in direct discourse (\cjRL{mh}, \cjRL{`th}, \cjRL{h}, \cjRL{ky}, \cjRL{gM}, \cjRL{hnh}, \cjRL{'l}, \cjRL{n'}, \cjRL{'M}, \cjRL{l'}).
The term "man" (\cjRL{'y+s}) can be used in many ways: man, husband, human; someone. Its use and expressions using it are significantly more frequent in non-P texts (42P/213nonP). This may be an evolution of the language rather than a deliberate or theological change on the part of P.
Finally, for the terms "to enter" (\cjRL{bw'}) or "to go" (\cjRL{hlK}) to be characteristic of non-P, this may reflect a stronger interest in place, in travel in the original texts probably composed to legitimize sanctuaries or as etiological narratives whereas these aspects are less marked in the P texts.

\subsection{Exodus} \label{app_disc_exodus}

\subsubsection{Semantics} \label{app_disc_exodus_semantics}

For the P-texts of Exodus, the algorithm has extracted the semantic features of the tabernacle construction in Exod 25-31; 35-40 but does not give features of the P-texts that would be found elsewhere.  We find in the features: the different names of the holy place, "the holy one", "the dwelling", "the tent of meeting"; the materials used for the construction, "acacia wood", "pure gold", "bronze", "linen", "blue, purple, crimson yarns", etc.; the spatialization, "around", "outside"; the dimensions, "length", "cubit", "five"; the components, "altar", "curtain", "ark", "utensils", "table", and YHWH's orders to Moses, "you shall make". Thus, the algorithm has a good understanding of the terms specific to the construction of the Tabernacle but is not susceptible to a more general understanding of the characteristics of P in Exodus.
The non-P features are more interesting. For example, the use of the word "people" (\cjRL{`M}) appears primarily in the non-P texts because the priestly redactors usually preferred to use the term "assembly" (\cjRL{`dh}). The word "I" in the long form (\cjRL{'nky}) is considered non-P because the word 'ny, the short form, appears in P texts. The expression "to YHWH" does appear 24 times in non-P texts, e.g., "to cry out to YHWH"/"to speak to YHWH"/"to turn to YHWH", whereas P avoids the expression. This is easily understandable by a desire to give YHWH the initiative in all interactions. In P, it is he who demands, commands, and speaks. There are few dialogues. As for the terms Egypt (\cjRL{m.sryM}) and Pharaoh (\cjRL{pr`h}), they are indeed quantitatively more frequent in the non-Priestly texts of Exodus (respectively 36P/139$^{\rm{nonP}}$ et 26P/89$^{\rm{nonP}}$) as in Genesis.

\subsubsection{Grammar} \label{app_disc_exodus_grammar}

As we have already seen, non-P texts more often adopt the protagonists' point of view by including dialogues or their thoughts, whereas P texts prefer a third-person narration. One of the consequences is the privileged use of 3rd singular or plural suffixes, unlike non-P texts where 1st singular or 2nd singular suffixes are more often used. Moreover, the massive use of the 3rd person in P texts can also be explained by the presence of pleonasms which use a form with this suffix: \cjRL{`mw}, \cjRL{'tw}, etc. \citep{holzinger1893einleitung}.
Concerning verbs, the form Qal or Piel, qatal in the 2nd masculine singular, is prevalent in P texts. This is understandable because of P’s theology, according to which God orders using the second person, and then the protagonists carry out according to YHWH's order. On the side of the non-P texts, the narrative form, i.e., Qal, wayyiqtol in the 3rd masculine singular, is significant, although these forms are  also very present in P texts.
Another peculiarity is the use in non-P of "name in the constructed form + place name". P seems to have avoided topical constructions because of reduced interest in localizations.
The remaining terms are persistent elements. Further analysis would be needed to understand the relevance of the distinction made by the algorithm.

\subsection{Summary} \label{app_disc_summary}

As we can see, the pipeline could extract typical features of priestly texts in Genesis, easily recognizable for a specialist. In addition, other P features have also been found that may be specific to a single narrative, correspond to repeated use of an expression, or be a significant theological theme such as water. On the other hand, the features of non-P texts do not indicate a coherent editorial milieu or style but rather allow us to better distinguish between P texts and non-P texts by particular theological or linguistic features.
The data provided by the algorithm allow for the detection of particularities that require an explanation. More detailed investigations than those presented above are necessary to better understand specific instances of the results. For the texts of Exodus, the excessive importance of the chapters devoted to the construction of the Tabernacle (Exod 25-31; 35-40) did not allow us to obtain satisfactory results in the characterization of P, which could indicate an originally independent document. Nevertheless, the characterization of non-P texts is relevant, as well as the results on grammar.

\end{document}